\documentclass[lettersize,journal]{IEEEtran}
\usepackage{amsmath,amsfonts}
\usepackage{algorithmic}
\usepackage{array}
\usepackage{textcomp}
\usepackage{stfloats}
\usepackage{url}
\usepackage{verbatim}
\usepackage{graphicx}
\usepackage{soul}
\usepackage{color}
\usepackage{ulem}
\usepackage{tabularx}
\usepackage{graphicx}
\usepackage{subfigure}

\usepackage{svg}

\usepackage{booktabs}
\usepackage{multirow}
\usepackage{tabularx}
\usepackage{graphicx}
\usepackage{svg}
\usepackage{algorithm}
\usepackage{algorithmic}
\usepackage{graphicx}

\def\BibTeX{{\rm B\kern-.05em{\sc i\kern-.025em b}\kern-.08em
    T\kern-.1667em\lower.7ex\hbox{E}\kern-.125emX}}
\usepackage{balance}
\begin{document}
\title{ARISE: Graph Anomaly Detection on Attributed Networks via Substructure Awareness}
\author{Jingcan~Duan,~Bin~Xiao$^{\dagger}$,~Siwei~Wang,~Haifang~Zhou,~Xinwang~Liu$^{\dagger}$,~\IEEEmembership{Senior~Member,~IEEE}
\thanks{Jingcan Duan, Haifang Zhou and Xinwang Liu are with the College of Computer Science and Technology, National University of Defense Technology, Changsha 410073, China. (E-mail: jingcan\_can@163.com, \{haifang\_zhou,\, xinwangliu\} @nudt.edu.cn).

Bin Xiao is with the Department of Computer Science and Technology, Chongqing University of Posts and Telecommunications, Chongqing 400065, China. (E-mail: xiaobin@cqupt.edu.cn).

Siwei Wang is with the Intelligent Game and Decision Laboratory, Beijing 100071, China (E-mail: wangsiwei13@nudt.edu.cn).

\IEEEcompsocthanksitem $^{\dagger}$: Corresponding author.}}

\markboth{IEEE Transactions on Neural Networks and Learning Systems}%
{How to Use the IEEEtran \LaTeX \ Templates}

\maketitle

\begin{abstract}
\textcolor{black}{Recently, graph anomaly detection on attributed networks has attracted growing attention in data mining and machine learning communities. Apart from attribute anomalies, graph anomaly detection also aims at suspicious topological-abnormal nodes that exhibit collective anomalous behavior. Closely connected uncorrelated node groups form uncommonly dense substructures in the network. However, existing methods overlook that the topology anomaly detection performance can be improved by recognizing such a collective pattern. To this end, we propose a new graph anomaly detection framework on attributed networks via substructure awareness (ARISE for abbreviation). Unlike previous algorithms, we focus on the substructures in the graph to discern abnormalities. Specifically, we establish a region proposal module to discover high-density substructures in the network as suspicious regions. The average node-pair similarity can be regarded as the topology anomaly degree of nodes within substructures. Generally, the lower the similarity, the higher the probability that internal nodes are topology anomalies.} To distill better embeddings of node attributes, we further introduce a graph contrastive learning scheme, which observes attribute anomalies in the meantime. In this way, \textcolor{black}{ARISE} can detect both topology and attribute anomalies. Ultimately, extensive experiments on benchmark datasets show that \textcolor{black}{ARISE} greatly improves detection performance (up to 7.30\% AUC and 17.46\% AUPRC gains) compared to state-of-the-art attributed networks anomaly detection (ANAD) algorithms.
\end{abstract}


\section{Introduction}
\IEEEPARstart{A}TTRIBUTED networks have the powerful capability of modeling and analyzing numerous complex scenarios in the real world. In recent years, researches~\cite{wu2020comprehensive} on attributed networks tasks are becoming an increasingly appealing direction for academia and industry. In particular, anomaly detection has been a vital topic in attributed networks~\cite{ma2021comprehensive}, which has ubiquitous applications in many fields, such as \textcolor{black}{financial fraud detection~\cite{zhang2022efraudcom}}, social spam detection~\cite{rao2021review}, network intrusion detection~\cite{mongiovi2013netspot}, and sinformation system error detection~\cite{boshmaf2013graph}.

\begin{figure}[!t]
\centering
\includegraphics[width = 0.48\textwidth]{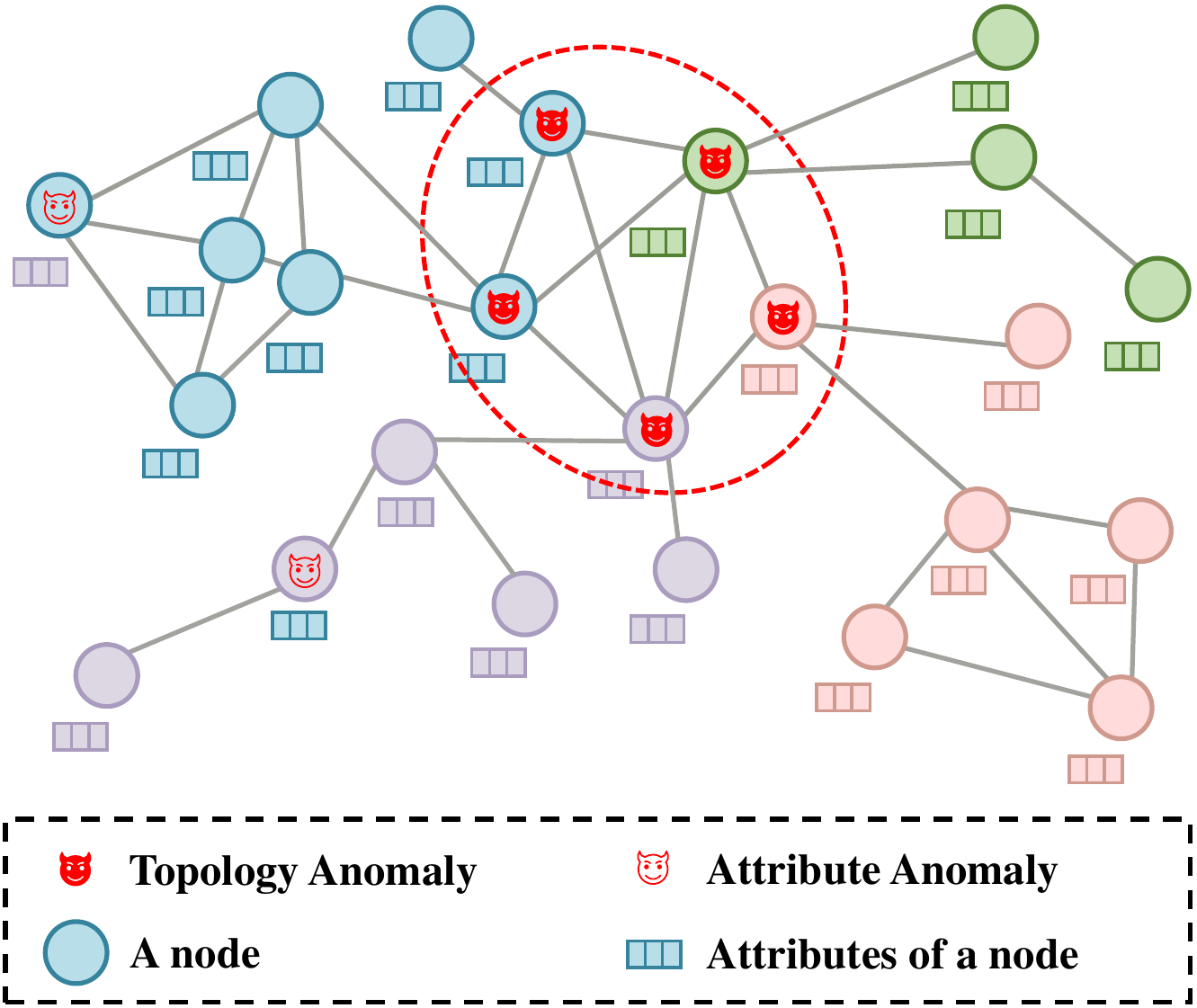}
% \vspace{5pt}
\caption{\textcolor{black}{The toy example contains two typical anomalous nodes in attributed networks: (1) Topology anomaly: uncorrelated nodes with dense links in the red circle. (2) Attribute anomaly: nodes with dissimilar attribute values from neighborhoods.}}
\label{fig:toy}  
\end{figure}

\textcolor{black}{Attributed networks anomaly detection (ANAD) intends to discern anomalous nodes that diverge from frequent patterns~\cite{ma2021comprehensive}.} Unlike the raw data widely applied in other fields~\cite{cheng2021improved, cheng2021unsupervised, hu2022detecting}, attributed networks include both network topology and node attribute information. \textcolor{black}{According to the previous literature~\cite{skillicorn2007detecting, song2007conditional, ding2019interactive, liu2022bond}, the mismatch between the above two types of information will cause two typical kinds of anomalies in attributed networks, i.e., the topology anomalies and attribute anomalies. As illustrated in Figure~\ref{fig:toy}, the topology anomaly has normal attribute values at a glance, while it densely links to the uncorrelated nodes~\cite{liu2021anomaly}; on the contrary, the attribute counterpart has a normal neighborhood connectivity relationship, while its attribute values may be contaminated by noise and are not similar to the neighborhoods. Further analysis demonstrates that the pattern of topology anomalies is the collective behavior of a group of anomalous nodes, which form uncommonly dense substructures in the networks.}

\textcolor{black}{Early shallow methods~\cite{breunig2000lof, xu2007scan, perozzi2016scalable, li2017radar, peng2018anomalous} primarily focus on feature engineering or pattern recognition methods to detect the anomalous behaviors of nodes. Such models rely on specific domain knowledge of researchers to improve detection performance. Besides, they cannot effectively dig deep non-linear information from the topology structure and the node attributes. The above reasons hinder them from discriminating abnormal nodes effectively.}

\textcolor{black}{Recently, the graph neural network (GNN) has made remarkable advances in many fields, e.g., natural language processing~\cite{marcheggiani2017encoding}, traffic systems~\cite{zhang2018gaan}, recommender systems~\cite{ying2018graph}, and chemistry~\cite{gilmer2017neural}. Therefore, GNN has become a natural choice for the ANAD task and can solve the problems faced by traditional methods better. DOMINANT~\cite{ding2019deep} designs a graph convolution network (GCN) based auto-encoder, which regards the reconstruction errors of structure and attribute matrices as the anomalous degrees. AnomalyDAE~\cite{fan2020anomalydae} replaces GCN with graph attention network (GAT). ResGCN~\cite{pei2021resgcn} improves GAT by adding residual information. Differently, \cite{tang2022rethinking} combines multiple graph filters and solves the anomalous information smoothing of nodes caused by GCN's low-pass filter. To avoid the labor-intensive and time-consuming cost of collecting node labels, CoLA first introduces the graph contrastive learning paradigm into ANAD. \cite{jin2021anemone, zheng2021generative, zhang2022reconstruction} boost the detection performance with the addition of node-level contrastive network, generative loss, and graph multi-view contrastive paradigm.}

\textcolor{black}{Although existing works have achieved impressive results in the detection performance, there is still a relative lack of research on effectively detecting topology anomalies. The common practice of most contrastive-based previous approaches is the comparison of attributes between the target node and its neighbors. Besides, reconstruction-based methods compare the original attribute and topology information of the target node with the one reconstructed from its neighbors. It is obvious that such a strategy is effective in detecting the anomalous behaviors of attribute anomalies. An attribute anomalous node can be easily captured if it is dissimilar from its neighbors. However, the topology anomaly has normal attribute values but more complex connectivity relationships with other topology anomalies. It is difficult to discern them under such a strategy. Therefore, a new algorithm that can effectively detect topology anomalies and cooperate well with attribute anomaly detection needs to be explored.}

\textcolor{black}{In this paper, we make a new attempt to remedy the mentioned defects in ANAD. To this end, we propose a novel graph \textbf{A}nomaly detection framework on att\textbf{RI}buted networks via \textbf{S}ubstructure awar\textbf{E}ness (termed \textbf{ARISE}), which both detects the topology and attribute anomalies concurrently. Topology anomalies are groups of uncorrelated nodes with dense links between them. Such a group of anomalous nodes viewed as a whole will form an uncommonly dense substructure in the networks. It is the anomalous behavior manifested by nodes combined together. However, existing works are unaware that the key to topology anomaly detection is discerning such a collective pattern from the network topology. To overcome this challenge, we elaborate a new topology detection scheme from a substructure-based perspective, which can detect their collective anomalous behaviors. To be specific, we first leverage a density-based algorithm in the region proposal module to discover high-density substructures as suspicious regions. Then, the anomalous degrees of the internal nodes are measured by their attribute similarities. On the other hand, we adopt a graph contrastive network for attribute anomaly detection. It can distill better node embeddings for similarity measures in topology anomaly detection simultaneously. Finally, a well-designed aggregator is utilized to fuse the topology and attribute anomalous information. Extensive evaluations and comprehensive metrics on benchmarks show ARISE's astonishing effectiveness compared to strong competitors.} 

In general, the main contributions of this work are summarized below:

\begin{itemize}
\item \textcolor{black}{We redesign a new principle to detect the topology anomalies. Specifically, it perceives their collective anomalous behaviors from the perspective of network substructures.}
\item \textcolor{black}{We propose a new integrated anomaly detection framework (ARISE) in attributed networks, which experts in detecting topology anomalies and attribute anomalies simultaneously.}
\item \textcolor{black}{Experiments demonstrate the remarkable advantage of ARISE against current attributed networks anomaly detection methods.} 
\end{itemize}

\begin{figure*}[!t]
    \centering
    \includegraphics[width = \textwidth]{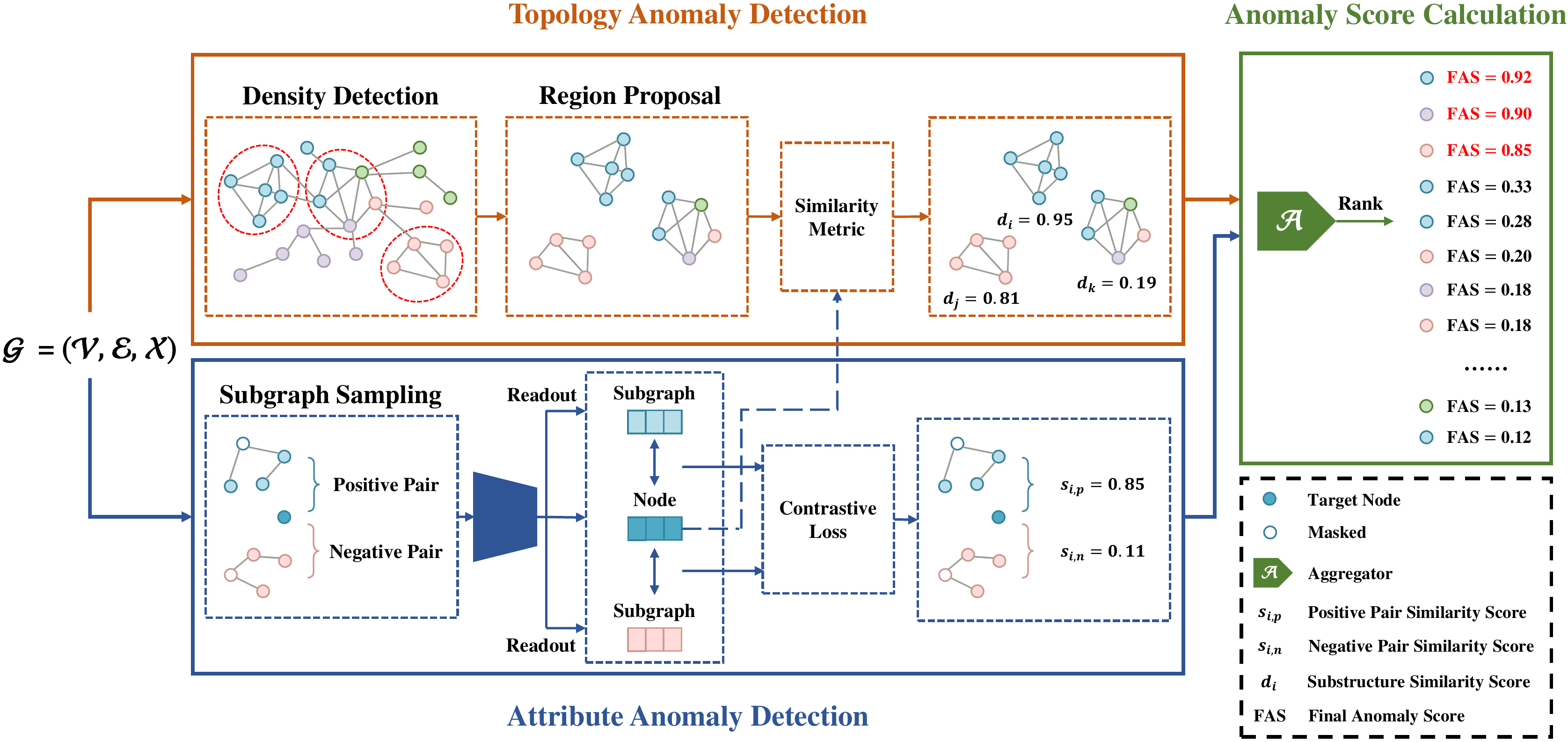}
    \caption{Overview of the \textcolor{black}{ARISE} model. It consists of three modules: (1) Topology anomaly detection: We design a region proposal module to perceive high-density substructures as suspicious regions. \textcolor{black}{The average node-pair similarity indicates the anomaly degrees of nodes within the substructure;} (2) Attribute anomaly detection: The target node forms a positive pair with its located subgraph and forms a negative pair with the irrelevant subgraph. After the contrastive learning network, the node embeddings are also used in topology anomaly detection; (3) Anomaly score calculation: \textcolor{black}{We treat the reciprocal of average node-pair similarity as the topology anomaly estimates of its inner nodes and adopt a multi-scale substructure scoring strategy. The node attribute anomaly score is calculated by measuring the relationship between it and its contrastive subgraphs.} Finally, the aggregator calculates the final anomaly score by fusing the attribute and topology anomaly scores of each node.}
    \label{fig:model}
\end{figure*}

\section{Related Work}\label{relatedWork}
In this section, we will review the related work in terms of (1) attributed networks anomaly detection; and (2) graph contrastive learning.

\subsection{Attributed Networks Anomaly Detection}
An attributed network is a structured network with multi-dimensional attribute information. Many traditional algorithms have been proposed in ANAD. They are based on the non-deep anomaly detection way, which captures anomalous patterns from either topology or node attributes of the network. LOF~\cite{breunig2000lof} and Radar~\cite{li2017radar} find the differences in attributes between the target node and the context or most of the nodes. Among the methods based on topology information, SCAN~\cite{xu2007scan} is one of the most well-known works. AMEN~\cite{perozzi2016scalable} and ANOMALOUS~\cite{peng2018anomalous} both use node attribute information and network topology information to achieve powerful results. However, the traditional methods cannot excavate deep information, making it difficult to improve performance continuously.

The emergence of GNN has dramatically improved the ability of the algorithms to extract the deep information of the attributed networks. They have been widely used in many fields~\cite{he2020lightgcn, chen2020revisiting, wu2018socialgcn, liu2019deep, shuai2022review}. It is natural to introduce GNN to graph anomaly detection.  AAGNN~\cite{zhou2021subtractive}, which is improved from the traditional anomaly detection algorithm, incorporates the one-class SVM algorithm into attributed networks to identify anomalous samples. Algorithms based on the self-supervised learning paradigm have also achieved great success. The generative self-supervised learning method DOMINANT~\cite{ding2019deep} performs anomaly detection by calculating the differences between the attribute matrix, the adjacency matrix, and the original matrices after GCN reconstruction. \textcolor{black}{COMMANDER~\cite{ding2021cross} pays attention to solving the cross-domain problem in ANAD.} Auxiliary attribute-based HCM~\cite{huang2021hop} adopts a node-to-hop count pretext task that takes the predicted values of hop counts of nodes and first-order neighbors as node anomaly scores. CoLA~\cite{liu2021anomaly} introduces the contrastive learning paradigm to ANAD for the first time, constructing node- and subgraph-level comparisons to identify anomalies by the differences between the node and contrastive subgraph attributes in positive and negative pairs. ANEMONE~\cite{jin2021anemone}, SL-GAD~\cite{zheng2021generative}, and Sub-CR~\cite{zhang2022reconstruction} do further research based on CoLA. Multi-scale strategies~\cite{wang2022progressive} have been proven effective in other domains. ANEMONE forms multi-scale contrastive with new node-level comparison and captures more anomalous information. SL-GAD adopts both contrastive learning and reconstruction patterns. Sub-CR works in a multi-view way with a global graph view generated by the diffusion model. \textcolor{black}{Differently, PAMFUL~\cite{zhao2021synergistic} captures global information by a synergistic approach.} However, most of these methods focus on digging the information of node attributes and pay less attention to the topological information.

\section{Graph Contrastive Learning}
\label{GCL}
The primary purpose of graph neural network algorithms is to transform the attributes of nodes into low dimensions according to the needs of specific tasks. The expensive label-collection progress limits the supervised and semi-supervised learning algorithms. As an essential branch of the graph self-supervised paradigm, graph contrastive learning can compensate for the above problem to a certain extent by obtaining enough supervision information from ingeniously designed pretext tasks. Based on the contrastive scale~\cite{liu2022graph}, most graph contrastive learning methods can be divided into two leading mainstream. One is same-scale contrastive learning. GRACE~\cite{zhu2020deep} generates two comparison views by graph augmentation methods that bring the positive pair nodes in the view closer and push the negative pair nodes farther. After two graph views are generated, HeCO~\cite{wang2021self} is trained on the heterogeneous graph with encoders maximizing the mutual information of the same nodes in the views. To alleviate the burden of constructing negative pairs, BGRL~\cite{thakoor2021bootstrapped} uses the Siamese network approach for model training. The other one is cross-scale contrastive learning. The pioneering work DGI~\cite{velickovic2019deep} proposes to maximize the mutual information of node representation and graph level representation to obtain global information. After generating two views through data augmentation, MVGRL~\cite{hassani2020contrastive} maximizes the node representation in one view and the graph level representation in the other view. Therefore, we focus on using cross-scale contrastive learning, which provides a better solution to distill the representation of node attributes for ANAD tasks.

\section{The Proposed Method}
In this section, we will first define the main problem of ANAD formally. The primary notations used throughout this paper are summarized in Table \ref{table:symbol}. Then, we introduce the proposed framework, \textcolor{black}{ARISE}. \textcolor{black}{Finally, we conduct complexity analysis of the model.}

\subsection{Attributed Networks Anomaly Detection}
\textcolor{black}{An attributed network $\mathcal{G} = \left ( \mathcal{V}, \mathcal{E}, \mathbf{X} \right ) $ consists of: (1) the set of nodes $\mathcal{V}$, where $\left |\mathcal{V}\right | = n$; (2) the set of edges $\mathcal{E}$, where $\left |\mathcal{E}\right | = m$; (3) the node attributes matrix $\mathbf{X} \in \mathbb{R}^{n\times d}$, indicates the attribute values of nodes. In addition, the adjacency matrix $\mathbf{A} \in \mathbb{R}^{n\times n}$ contains the topology information of the network, and $\mathbf{A}_{ij} = 1$ represents there is an edge between node $v_{i}$ and $v_{j}$, otherwise, $\mathbf{A}_{ij} = 0$. In attributed networks anomaly detection, the model is trained to learn a function $score\left(v_{i}\right)$ that calculates the anomaly score for node $v_{i}$ in $\mathcal{G}$. The larger the anomaly score, the more likely the node is anomalous. It is worth noting that we conduct the attributed networks anomaly detection task strictly unsupervised in this paper.}

\begin{table}[t]
\centering
\caption{Notation summary}
\begin{tabularx}{0.45\textwidth}{p{2.0cm}<{\centering}p{5.5cm}}
\toprule
$\textbf{Notations}$ & $\textbf{Definitions}$\\
\midrule
$\mathcal{G}$       & An attributed network\\
$\mathcal{C}_j$       & \textcolor{black}{A substructure in the network}\\
$v_{i}$       & \textcolor{black}{The $i$-th node of $\mathcal{G}$}\\
\textcolor{black}{$\mathbf{A}\in\mathbb{R}^{n\times n}$}       & Adjacency matrix of $\mathcal{G}$\\
\textcolor{black}{$\mathbf{D}\in\mathbb{R}^{n\times n}$} &Degree matrix of $\mathbf{A}$\\
\textcolor{black}{$\mathbf{X}\in\mathbb{R}^{n\times d}$}       & Attribute matrix of $\mathcal{G}$\\
\textcolor{black}{$\boldsymbol{x_{i}}\in\mathbb{R}^{1\times d}$}    & Attribute vector of $v_{i}$\\
\textcolor{black}{$\mathbf{H}^{\left ( \ell \right ) }\in\mathbb{R}^{n\times d^{\prime}}$} &Hidden subgraph representation of the ${\ell}$-th layer\\
\textcolor{black}{$\boldsymbol{h}^{\left ( \ell \right ) }\in\mathbb{R}^{1\times d^{\prime}}$} &\textcolor{black}{Node hidden representation of the ${\ell}$-th layer}\\
\textcolor{black}{$\mathbf{W}^{\left ( \ell \right )}\in\mathbb{R}^{d^{\prime} \times d^{\prime}}$} &Network parameters of the ${\ell}$-th layer\\
$R^{t}$&Number of topology anomaly detection round\\
$R^{a}$&Number of attribute anomaly detection round\\
% $s_{i,p}^{a}$&The similarity between $v_{i}$ and subgraph in a positive pair\\
\textcolor{black}{$y_{i}\in \left \{ 0,1 \right \}$} & \textcolor{black}{Ground-truth label of a instance pair}\\
\textcolor{black}{$score\left(v_{i}\right)$}&Final anomaly score of $v_{i}$\\
\bottomrule
\end{tabularx}
\label{table:symbol}
\end{table}

\subsection{Proposed Method}
In this subsection, we will introduce our method \textcolor{black}{ARISE}. As shown in Figure~\ref{fig:model}, it consists of three main components. The first part is the topology anomaly detection module, which uses a density-based algorithm to perceive high-density substructures as suspicious regions. \textcolor{black}{The anomaly degrees of internal nodes are evaluated by the average node-pair similarity in the substructure.} The second part is the attribute anomaly detection module, which forms a contrast of node-subgraph level to obtain node attribute anomaly degree. It should be pointed out that these two modules leverage both node attribute information and network topology information. The last one is the anomaly score calculation, which aggregates the anomalous information from the first two modules to obtain the final anomaly score for each node. Such two types of anomaly score calculations are performed under multi-round detection. \textcolor{black}{In each detection, the former employs a multi-size substructure scoring strategy, which takes into account the size of the substructure.} The latter combines the effects of both positive and negative pairs.

\subsubsection{Topology Anomaly Detection}
\textcolor{black}{Previous methods~\cite{liu2021anomaly, jin2021anemone, zheng2021generative} constitute node-subgraph pairs in the contrastive network by random walk~\cite{tong2006fast, perozzi2014deepwalk, qiu2020gcc, wang2015unsupervised, wang2013towards}. Nevertheless, the uncertainty caused by random walk does not always contribute to learning the node-local topology information that is beneficial for the task. In extreme cases, the sampled subgraph around an anomalous node may contain a few similar nodes in its neighborhoods, which will harm detecting it. On the other hand, this node-subgraph contrastive mode ignores the collective abnormal behavior of topology anomalies on the network topology. Therefore, we make a new attempt to leverage local topology information from the perspective of substructures in the networks. We design a region proposal module to extract suspicious substructures and overcome the above deficiency. Afterward, we discriminate the anomalous nodes in the substructure based on the average node-pair similarity.}

\textbf{Region Proposal Module.} \textcolor{black}{A group of topology anomalies will form an uncommonly dense substructure in the networks. To this end, we employ a density-based algorithm to discover the high-density substructures that may contain topology anomalies. In practice, we first utilize a $k$-core method~\cite{batagelj2003m} to find the $k$-core graphs. Then they will be partitioned into substructures based on the node connectivity.} In addition to this straightforward density-based algorithm, other more sophisticated algorithm strategies are also applicable under our substructure-aware framework as one of the potential directions in the future. Ultimately, these substructures $\left \{\mathcal{C}_{0},..., \mathcal{C}_{b}\right \}$ with possible anomalous nodes can be regarded as suspicious regions.

\textcolor{black}{\textbf{Topology Anomaly Degree Estimate.} In reality, there are some scenarios where normal samples possess close connections between them, e.g., there are frequent email correspondences between managers and their secretaries. Conversely, the intensive email exchanges between assembly line workers and secretaries could be regarded as abnormality~\cite{yu2016survey}. Therefore,} some suspicious regions have only normal nodes, while others contain normal nodes and topology anomalies. \textcolor{black}{In light of this, we measure the anomaly degree of internal nodes by average node-pair similarity in the substructure. It should be pointed out that the nodes with more significant differences in attributes have a higher anomalous possibility.} To obtain more accurate similarity, we utilize the embeddings of node attributes instead of the original attribute values. \textcolor{black}{Specifically, we choose cosine distance as the node-pair similarity measurement:}

\begin{equation}
d_{k,q}=Similarity\left (\boldsymbol{z}_{k},\boldsymbol{z}_{q} \right )=\cos \left (\boldsymbol{z}_{k},\boldsymbol{z}_{q}\right ),
\label{cos}
\end{equation}

\noindent where $\boldsymbol{z}_{k}$ and $\boldsymbol{z}_{q}$ represents the embedding of node $v_{k}$ and $v_{q}$.

\textcolor{black}{After that, we utilize an average function to compute the average node-pair similarity for each substructure:}

\textcolor{black}{
\begin{equation}
d_{j}=\frac{1}{\hat{n}_{j}}\sum_{l=1}^{\hat{n}_{j}}d_{k,q}^{\left (l\right )},
\label{sim_comm}
\end{equation}
where $\hat{n}_{j}$ is the number of the node pairs in the substructure $\mathcal{C}_{j}$. And node $v_{k}$ and $v_{q}$ belong to $\mathcal{C}_{j}$.}

So far, it has not been solved how to obtain more qualified node embeddings for the similarity metric in Eq.\eqref{cos}. In addition, it is essential to detect attribute anomalies in ANAD. Hence, we introduce the graph contrastive learning scheme in attribute anomaly detection, which can acquire better node representations in the meantime.

\subsubsection{Attribute Anomaly Detection}
The recent success of graph contrastive learning introduced in Section~\ref{GCL} has shown great potential to discern attribute anomalies. Inspired by this, we define contrastive instance pairs following~\cite{liu2021anomaly}. Afterward, the node attribute anomaly degree is calculated by measuring the similarity between the embeddings of instance pairs. More importantly, \textcolor{black}{ARISE} can learn better node representations served in the topology anomaly detection module.

\textbf{Subgraph Sampling.} The construction of positive and negative pairs is vital for contrastive learning. While there are multiple-scale contrasts, one of the most potent comparisons is the contrast of node and subgraph for ANAD. The supervision information from the subgraph includes both neighborhood attributes and topology. For the target node $v_{i}$, it forms a positive pair with the subgraph where it is located and forms a negative pair with the subgraph where another random node is located. To avoid the interaction between the node and subgraphs, the attribute values of the node are masked, i.e., all attribute values are set to 0.

There are many subgraph sampling methods, but we use a random walk-based algorithm in this paper. Because it is practical and straightforward to implement. A generated subgraph is composed of a node and its neighborhoods.

\textbf{Contrastive Learning Network.} The contrastive learning network pulls the positive pairs and pushes the negative pairs into the latent space. In this process, the node representations are optimized. Firstly, we learn the representation of the subgraph \textcolor{black}{corresponding to the target node $v_{i}$} through a GCN layer. The hidden layer representation of the subgraph in the positive pair can be defined as:

\textcolor{black}{
\begin{equation}
\mathbf{H}^{\left (\ell+1  \right ) }_{i} =\sigma \left ( \mathbf{\widetilde{\mathbf{D}}}^{-\frac{1}{2}}_{i}\mathbf{\widetilde{\mathbf{A}}}_{i}\mathbf{\widetilde{\mathbf{D}}}^{-\frac{1}{2}}_{i}\mathbf{H}^{\left (\ell  \right ) }_{i}\mathbf{W}^{\left ( \ell \right ) }\right ),
\end{equation}where $\mathbf{H}^{\left (\ell+1  \right ) }_{i}$ and $\mathbf{H}^{\left (\ell  \right ) }_{i}$ denote the hidden representation of the ${\left (\ell+1\right ) }$-th and ${\ell}$-th layer, $\mathbf{\widetilde{\mathbf{D}}}^{-\frac{1}{2}}_{i}\mathbf{\widetilde{\mathbf{A}}}_{i}\mathbf{\widetilde{\mathbf{D}}}^{-\frac{1}{2}}_{i}$ is the normalization of the adjacency matrix, $\mathbf{W}^{\left (\ell \right ) }$ denotes the network parameters, $\sigma \left (\cdot   \right )$ is activation function ReLU here.}

After GCN, we manage to define a \textit{Readout} function, which converts the embedding of the subgraph \textcolor{black}{$\mathbf{E}_{i}$} to the same shape as \textcolor{black}{$v_{i}$}. In practice, we choose average to achieve \textit{Readout}:

\textcolor{black}{
\begin{equation}
\boldsymbol{e}_{i}=Readout\left (\mathbf{E}_{i}\right ) =\frac{1}{\tilde{n}_{i}}\sum_{l=1}^{\tilde{n}_{i}} \left (\mathbf{E}_{i} \right )_{l},
\end{equation}
where $\tilde{n}_{i}$ denotes the number of nodes within the subgraph, $\left (\mathbf{E}_{i} \right )_{l}$ denotes the ${\ell}$-th row of $\mathbf{E}_{i}$, and $\boldsymbol{e}_{i}$ is the final representation of the subgraph.}

Then, we need to encode the attribute of the target node into the same embedding space as subgraphs. The embedding of the target node can be acquired by using the model parameters of GCN:

\begin{equation}
\boldsymbol{h}^{\left (\ell+1 \right ) }_{i} =\sigma \left (\boldsymbol{h}^{\left (\ell \right ) }_{i}\mathbf{W}^{\left ( \ell \right ) }\right ),
\end{equation}
where $\boldsymbol{h}^{\left (\ell+1 \right ) }_{i}$ and $\boldsymbol{h}^{\left (\ell \right ) }_{i}$ denote the hidden representation of the ${\left (\ell+1\right ) }$-th and ${\ell}$-th layer, $\mathbf{W}^{\left ( \ell \right ) }$ denotes the network parameters, $\sigma \left (\cdot \right )$ is ReLU here. The last two are both shared with GCN. $\boldsymbol{z}_{i}$ is the final representation of the target node.

\textcolor{black}{We utilize a bilinear model to tease out the relationship between the target node $v_{i}$ and its counterpart subgraph, i.e., the similarity:}

\textcolor{black}{
\begin{equation}
s_{i} =Bilinear\left (\boldsymbol{z}_{i},\boldsymbol{e}_{i}\right ) =sigmoid \left (\boldsymbol{z}_{i}\widetilde{\mathbf{W}}\boldsymbol{e}_{i}^\top\right ),
\label{sample}
\end{equation}
where $\widetilde{\mathbf{W}}$ denotes the parameter matrix, and ${sigmoid} \left (\cdot  \right )$ is the logistic sigmoid function.}

According to the characteristics of the ANAD task, we decide to apply the binary cross-entropy loss~\cite{velickovic2019deep} as the objective function of the module. It can be expressed as follows:
\textcolor{black}{
\begin{equation}
\mathcal{L}=- \sum_{i=1}^{n}\left (y_{i}\log{\left (s_{i} \right )} +  \left (1 - y_{i}\right )\log{\left ( 1 - s_{i} \right ) }\right ), 
\label{loss}
\end{equation}
where $y_{i}$ is equal to 1 in positive pairs, while is equal to 0 in negative pairs.}

\subsubsection{Anomaly Score Calculation} \textbf{Topology Anomaly Score.} \textcolor{black}{Since the average node-pair similarity is negatively correlated with the anomaly degree of nodes in the substructure, the reciprocal of the average node-pair similarity is naturally used to represent their anomaly degrees. Then we treat the reciprocal value as the topology anomaly estimate of each internal node within the substructure.}
% If a node is not in any previously detected substructure, its topology anomaly score is 0. 
\textcolor{black}{We calculate the topology anomaly estimate for each node in the substructure:} 
\textcolor{black}{
\begin{equation}
t_{i}=\frac{1}{d_{j}},
\label{topo_score1}
\end{equation}
where $d_{j}$ is the inner-node similarity of substructure $\mathcal{C}_{j}$. And $t_{i}$ represents the topology anomaly score of the target node $v_{i}$ that belongs to $\mathcal{C}_{j}$.}

\textcolor{black}{It is unknown which substructure the topology anomalies belong to. We perform multi-round detections with a gradually increasing $k$. To avoid unnecessary calculations, we start the detections with a specific $k$ equal to the average node degree $\delta$. The substructure discovery algorithm will be stopped when returning NULL. In each detection, we compute the topology anomaly estimates for the nodes detected in any substructure via Eq.~\eqref{cos},~\eqref{sim_comm}, and~\eqref{topo_score1}. The other nodes that have not been detected in any substructure will be assigned with a topology anomaly score of 0.} \textcolor{black}{\cite{noble2003graph, he2012large} indicate that smaller substructures appear frequently in the network while larger substructures appear infrequently. According to the characteristics of ANAD, we design the topology anomaly scoring method by combining the average node-pair similarity and the substructure size. In general, we adopt a new multi-size substructure scoring strategy for topology anomaly detection.} In practice, we adopt the number of nodes in the substructure to measure this impact. Therefore, we finally average the anomaly scores from multi-round detection to determine the topology anomaly score for each node. So the final topology anomaly score of the target node can be defined as follows:

\textcolor{black}{
\begin{equation}
score\_t\left ( v_{i} \right )=\frac{1}{R^{t}}\sum_{r=1}^{R^{t}}\left | C_{j} \right |t_{i}^{\left (r\right )},
\label{topo_score2}
\end{equation}
where $\left | C_{j} \right |$ indicates the number of the node in the substructure $\mathcal{C}_{j}$ that $v_{i}$ belongs to. And $R^{t}$ is the number of topology anomaly detection rounds. $score\_t\left ( v_{i} \right )$ denotes the final topology anomaly score of the target node $v_{i}$.}

\textbf{Attribute Anomaly Score.} The attribute anomaly detection module obtains the similarities between the embeddings of each target node and contrastive subgraphs. We calculate the node attribute anomaly score according to the similarities information. For a normal node, its embedding should be similar to the subgraph in the positive pair, i.e., $s_{i,p}$ is close to 1. On the contrary, it is dissimilar to the subgraph in the negative pair, i.e., $s_{i,n}$ is close to 0. For an anomalous node, its embedding should be dissimilar to the subgraphs in both positive and negative pairs, i.e., $s_{i,p}$ and $s_{i,n}$ are both close to 0.
% In some extreme instances, the anomalous node is instead similar to the subgraph in the negative pair. 
\textcolor{black}{Therefore, we define the attribute anomaly estimate of the target node as follows:}

\textcolor{black}{
\begin{equation}
a_{i} = s_{i,n} - s_{i,p},
\label{attr_score1}
\end{equation}
where $s_{i,n}$ is the similar degree of $v_{i}$ and its corresponding subgraph in a negative pair, and $s_{i,p}$ is the similar degree of $v_{i}$ and its corresponding subgraph in a positive pair.}

Due to the randomness of the subgraph sampling method, it cannot reflect all the attribute information of the node neighborhoods in one sampling. Thus, we perform multi-round detection and construct multiple groups of positive and negative pairs. Then the final attribute anomaly score of the target node is expressed as follows:
\textcolor{black}{
\begin{equation}
score\_a\left ( v_{i} \right )=\frac{1}{R^{a}}\sum_{r=1}^{R^{a}} a_{i}^{\left (r\right )},
\label{attr_score2}
\end{equation}
where $score\_a\left ( v_{i} \right )$ denotes the final attribute anomaly score of the target node $v_{i}$, and $R^{a}$ is the number of attribute anomaly detection round.}

\textbf{Final Anomaly Score.} \textcolor{black}{After the topology and attribute anomaly scores for each node are determined, we employ an aggregator to fuse them.} Firstly, each kind of score is normalized to unify its magnitudes. Then the aggregator sums the two scores according to weight and obtains the final anomaly score $score\left ( v_{i} \right )$ for each node:
\textcolor{black}{
\begin{equation}
score\left ( v_{i} \right ) = \left (1-\alpha \right ) \cdot  score\_t\left ( v_{i} \right )+\alpha \cdot  score\_a\left ( v_{i} \right ),
\label{final_score}
\end{equation}}where ${\alpha \in \left [ 0,1 \right ] }$ is a trade-off parameter to balance the importance between two scores.

In general, the overall procedures of \textcolor{black}{ARISE} are shown in Algorithm~\ref{ALGORITHM}.
\begin{algorithm}[!t]
\small
\caption{The proposed \textcolor{black}{ARISE}.}
\label{ALGORITHM}
\flushleft{\textbf{Input}: An attributed network $\mathcal{G} = \left ( \mathcal{V}, \mathcal{E}, \textbf{X} \right )$; Number of training epochs $E$; Batch size $B$.} \\
\flushleft{\textbf{Output}: Anomaly score function $score\left(\cdot \right)$.} 
\begin{algorithmic}[1]
\STATE // Topology anomaly detection
\STATE Calculate the average degree $\delta$ of nodes in graph.
\FOR{$k=\delta$ to $k_{max}$}
\STATE Detect high-density substructures $\left \{\mathcal{C}_{0},..., \mathcal{C}_{b}\right \}$.
\FOR{${C}_{j}$}
\STATE Measure the inner-node embedding similarities of ${C}_{j}$ via Eq.\eqref{cos}.
\STATE Calculate the average node-pair similarity of ${C}_{j}$ via Eq.\eqref{sim_comm}.
\ENDFOR
\ENDFOR
\STATE // Attribute anomaly detection
\FOR{$e=1$ to $E$}
\STATE $\mathcal{V}$ is divided into batches with size $B$ by random.
\FOR{$v_{i} \in B$}
\STATE Form the positive and negative pairs by random walk sampling.
\STATE Estimate the similarity scores for the embeddings of the target node and two subgraphs in the positive and negative pair via Eq.\eqref{sample}.
\STATE Calculate the loss $\mathcal{L}$ via Eq.\eqref{loss}.
\STATE Back propagation and update trainable parameters.
\ENDFOR
\ENDFOR
\STATE // Anomaly score calculation
\STATE By multiple round detection, calculate the final anomaly score for each node via Eq.\eqref{topo_score1}, \eqref{topo_score2}, \eqref{attr_score1}, \eqref{attr_score2}, and \eqref{final_score}.
\end{algorithmic}
\end{algorithm}

\subsection{Complexity Analysis}
\label{time_complexity}
We analyze the time complexity of each component in ARISE. In the topology anomaly detection module, the time complexity of $k$-core method is $\mathcal{O}\left ( \max \left ( n, m \right ) \right ) $ ($n$ is the number of nodes in the graph, and $m$ represents the number of edges in the network). Therefore, $R^{t}$ rounds of detection need $\mathcal{O}\left ( \max \left ( n, m \right )R^{t} \right ) $. For the attribute anomaly detection module, each RWR subgraph sampling for all nodes is $\mathcal{O}\left ( c\delta n \right )$~($c$ is the number of nodes within the subgraphs, and $\delta$ is the mean degree of the network). The time complexity of GCN for all nodes is $\mathcal{O}\left ( \left (Kqd+Kcd^{2}\right )n \right )$ ($K$ is the number of layers, $d$ is the dimension of node attributes in hidden space, and $q$ is the number of edges in subgraphs). Hence, this module needs $\mathcal{O}\left ( n \left (c\delta+Kqd+Kcd^{2}\right ) \left (T+ R^{a} \right ) \right )$ ($T$ is the number of epochs in training, and $R^{a}$ is the round of attribute detection). The overall time complexity of the proposed model is $\mathcal{O}\left ( \max \left ( n, m \right )R^{t} + n \left (c\delta+Kqd+Kcd^{2}\right ) \left (T+ R^{a} \right ) \right )$.

\textcolor{black}{In the meantime, we conduct time complexity analysis of recent important ANAD methods. Table~\ref{table:time} shows that since the augmented view is generated by graph diffusion, Sub-CR~\cite{zhang2022reconstruction} has the largest time complexity of all the methods. In summary, our method improves performance without significantly increasing complexity.}

\begin{table}[ht]
\centering
\caption{Comparison of time complexity for different anomaly detection methods in attributed networks. $R$ is the number of rounds in inference for the compared methods. And the other notations are the same as subsection~\ref{time_complexity}.}
\resizebox{0.48\textwidth}{!}{
\begin{tabular}{ccc}
\toprule
\textbf{Method}& \textbf{Time Complexity} & \\
\midrule
LOF~\cite{breunig2000lof} &$\mathcal{O}\left (n\log{n} \right )$\\
ANOMALOUS~\cite{peng2018anomalous} &$\mathcal{O}\left (n^{2}d \right )$\\
DOMINANT~\cite{ding2019deep} &$\mathcal{O}\left ( \left (Kmd+Knd^{2} \right ) T \right )$\\
CoLA~\cite{liu2021anomaly} &$\mathcal{O}\left ( n \left (c\delta+Kqd+Kcd^{2}\right ) \left (T+ R \right ) \right )$\\
ANEMONE~\cite{jin2021anemone} &$\mathcal{O}\left ( n \left (c\delta+Kqd+Kcd^{2}\right ) \left (T+ R \right ) \right )$\\
SL-GAD~\cite{zheng2021generative} &$\mathcal{O}\left ( n \left (c\delta+Kqd+Kcd^{2}\right ) \left (T+ R \right ) \right )$\\
HCM~\cite{huang2021hop} &$\mathcal{O}\left ( m + \left ( m+n\right )\log{n} + KmdT \right )$\\
Sub-CR~\cite{zhang2022reconstruction} &$\mathcal{O}\left ( n^{3}+n \left (c\delta+Kqd+Kcd^{2}\right ) \left (T+ R \right ) \right )$\\
\midrule
Proposed &$\mathcal{O}\left ( \max \left ( n, m \right )R^{t} + n \left (c\delta+Kqd+Kcd^{2}\right ) \left (T+ R^{a} \right ) \right )$\\
\bottomrule 
% \vspace{-25pt}
\end{tabular}
}
\label{table:time}
\end{table}

\section{Experiment}
To verify the effectiveness of \textcolor{black}{ARISE}, in this section, we conduct experiments on six benchmark datasets and evaluate the model through three metrics.

\subsection{Experimental Settings}
\subsubsection{Datasets}
We perform experiments on six datasets commonly used in ANAD. The details of each dataset are presented in Table \ref{table:datasets}. They are Cora~\cite{sen2008collective}, CiteSeer~\cite{sen2008collective}, DBLP~\cite{yuan2021higher}, Citation~\cite{yuan2021higher}, ACM~\cite{yuan2021higher}, and PubMed~\cite{sen2008collective}. DBLP is an author network. There is an edge between two authors if they have a coauthor relationship. The others are publication network datasets, which are composed of scientific publications. Each paper is regarded as a node. The citation relations form the edges of the network.

\begin{table}[ht]
\centering
\caption{The statistics of datasets.}
\begin{tabular}{ccccclllll}
\toprule
$\textbf{Datasets}$&$\textbf{Nodes}$&$\textbf{Edges}$&$\textbf{Attributes}$&$\textbf{Anomalies}$\\
\midrule
$\textbf{Cora}$& 2708 & 5429 & 1433 & 150 \\
$\textbf{CiteSeer}$& 3327 & 4732 & 3703 & 150 \\
$\textbf{DBLP}$& 5484 & 8117 & 6775 & 300 \\
$\textbf{Citation}$& 8935 & 15098 & 6775 & 450 \\
$\textbf{ACM}$& 9360 & 15556 & 6775 & 450 \\
$\textbf{PubMed}$& 19717 & 44338 & 500 & 600 \\
\bottomrule
\end{tabular}
\label{table:datasets}
\end{table}

\subsubsection{Anomaly Injection}
\label{injection}
Since the above six datasets are commonly attributed network datasets, they are usually considered to have no anomalies. Therefore, we need to inject anomalous nodes into these datasets for the experiment to work successfully. Following the previous literature~\cite{ding2019interactive, skillicorn2007detecting, song2007conditional}, we inject both topology and attribute anomalies into the datasets. Considering the balance, such two types of anomalies have the same number in each dataset.

\textbf{Topology Anomaly Injection.} Topology anomalies are mainly generated by perturbing the topology of the attributed networks. If many nodes in the dataset with dissimilar attributes are much more closely connected than the average, this can be considered a typical anomalous situation~\cite{ding2019interactive, skillicorn2007detecting}. \textcolor{black}{Specifically, we randomly select $\hat{m}$ nodes that make them fully connected and repeat this operation $p$ times.} The number $\hat{m}$ is set to 15 in practice. \textcolor{black}{The number $p$ is 5, 5, 10,  15, 15, and 20 on Cora, CiteSeer, DBLP, Citation ACM, and PubMed.}

\textbf{Attribute Anomaly Injection.} Attribute anomalies are mainly generated by perturbing the attribute values of nodes. Using the method proposed in the literature~\cite{song2007conditional}, attribute anomalies are generated by modifying node attribute values. Firstly, we randomly select a node $v_{i}$ and another $\hat{n}$ nodes. The number $\hat{n}$ is usually 50. Then we calculate the Euclidean distance between $v_{i}$ and each of the $\hat{n}$ nodes. Finally, the attribute values of $v_{i}$ are changed to be the same as that of node $v_{j}$, which has the largest Euclidean distance. The number of attribute anomalies is 75, 75, 150, 225, 225, and 300 on each dataset, respectively.

\subsubsection{Baselines}
We select eight well-known ANAD algorithms to compare with our proposed framework, \textcolor{black}{ARISE}. They are LOF~\footnote{https://github.com/damjankuznar/pylof}~\cite{breunig2000lof}, ANOMALOUS~\footnote{https://github.com/zpeng27/ANOMALOUS}~\cite{peng2018anomalous}, DOMINANT~\footnote{https://github.com/kaize0409/GCN\_AnomalyDetection\_pytorch}~\cite{ding2019deep}, CoLA~\footnote{https://github.com/GRAND-Lab/CoLA}~\cite{liu2021anomaly}, ANEMONE~\footnote{https://github.com/GRAND-Lab/ANEMONE}~\cite{jin2021anemone}, SL-GAD~\footnote{https://github.com/KimMeen/SL-GAD}~\cite{zheng2021generative}, HCM~\footnote{https://github.com/Juintin/GraphAnomalyDetection}~\cite{huang2021hop} and Sub-CR~\footnote{https://github.com/Zjer12/Sub}~\cite{zhang2022reconstruction}. The first two are non-deep methods, and the last six are deep methods. Following the operation of CoLA, the datasets are first pre-processed before running ANOMALOUS. Specifically, the attributes dimension of nodes is reduced to 30 by PCA.

\subsubsection{Evaluation Metric}
Our experiments employ three common metrics to evaluate the model comprehensively.

\textbf{AUC.} The ROC curve is a plot of the true positive rate against the false positive rate according to the ground truth and the anomaly scores of the nodes. AUC value is the area under the ROC curve. A higher value means the score function assigns a higher score to the randomly chosen anomalous sample than a normal sample.

\textbf{AUPRC.} The PR curve is a plot of precision against recall according to the ground truth and the anomaly scores of the nodes. AUPRC value is the area under the PR curve, reflecting the performance of the positive samples. In practice, we use the method of calculating the average precision to obtain AUPRC value~\cite{schutze2008introduction}.

\textbf{Precision@K.} It indicates the proportion of true-positive samples among the top $k$ samples with the largest anomaly scores. The value of Precision@K ranges from 0 to 1, and close to 1 means the model performs better.

\subsubsection{Parameter Settings}
In the attribute anomaly detection module, we set the size of subgraphs in positive pairs and negative pairs to be 4. In the meantime, the basic network model (GCN) is one layer, and the latent embedding dimension is 64 on all datasets. The learning rate of the model for Cora, CiteSeer, DBLP, Citation, ACM, and PubMed is 0.003, 0.003, 0.0005, 0.001, 0.0005, and 0.001. To get better performance, the number of training epochs on Cora, CiteSeer, and PubMed is 100. It is set to 400 on DBLP, Citation, and ACM.

\subsubsection{Computing Infrastructures}
Except for ANOMALOUS, the experiments of other models run on the PyTorch 1.10.2 platform using an Intel (R) Core (TM) i9-10850K CPU (3.60GHz), 64GB RAM, and an NVIDIA GeForce RTX 3070 8GB GPU. The experiments of ANOMALOUS run on MATLAB R2020b platform using the same machine.

\subsection{Experimental Results}
\subsubsection{\textcolor{black}{Analysis of Performance Comparison}}
\begin{figure*}[t]
    \centering
    \includegraphics[width = \textwidth]{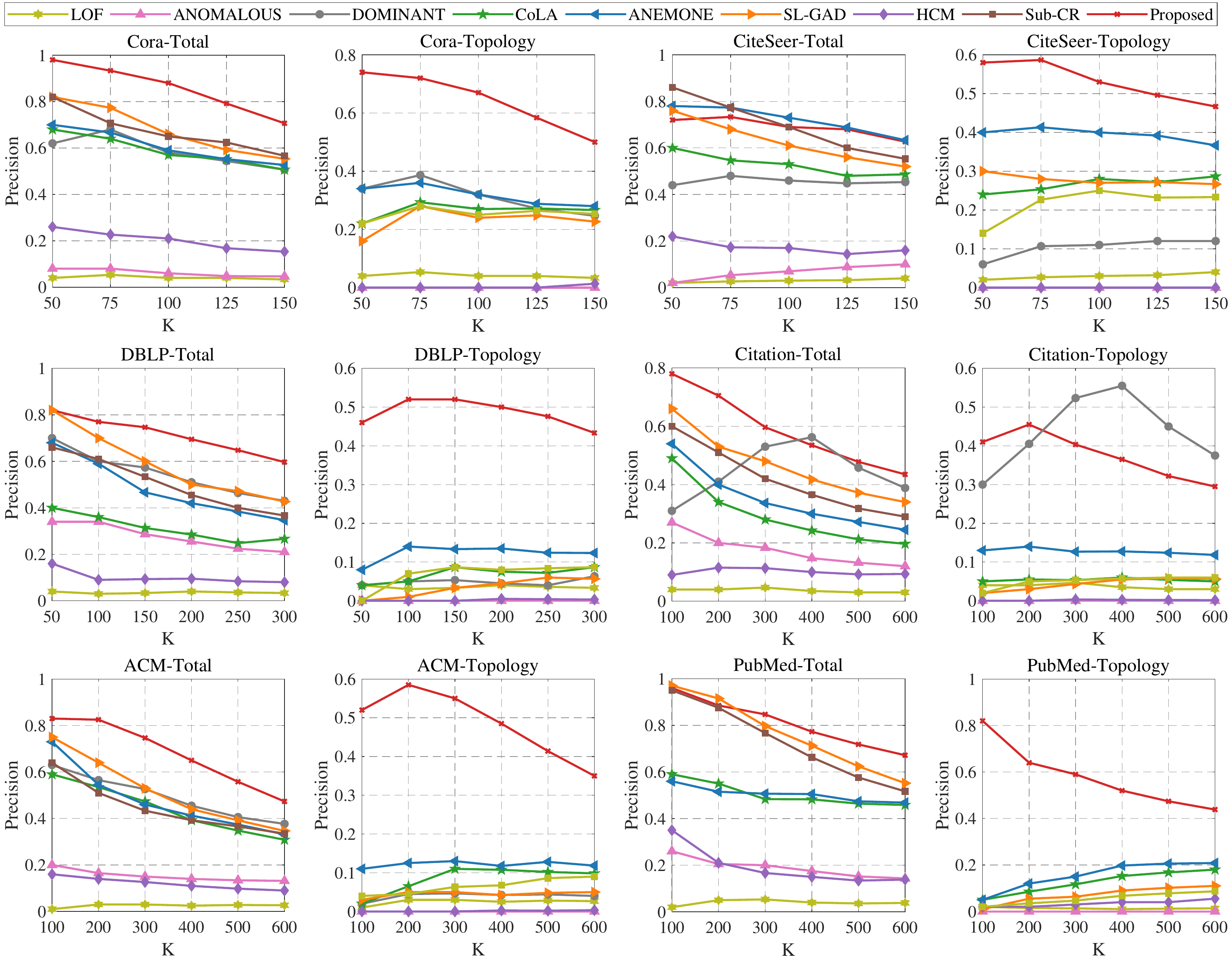}
    \caption{Performance comparison results w.r.t. Precision@K of total anomaly and topology anomaly on six benchmark datasets.}
    \label{fig:precision}
\end{figure*}

The ROC curves of these nine models on these six datasets are shown in Figure \ref{fig:AUC}. The specific AUC values and AUPRC values are shown in Table \ref{table:AUCAUPRC}. Meanwhile, the Precision@K of total anomaly and topology anomaly line diagrams are shown in Figure \ref{fig:precision}. Through comparison, we can conclude as follows:

\begin{figure*}[htbp]
\centering
\subfigure[ROC curve of Cora.]{
\includegraphics[width=0.31\textwidth]{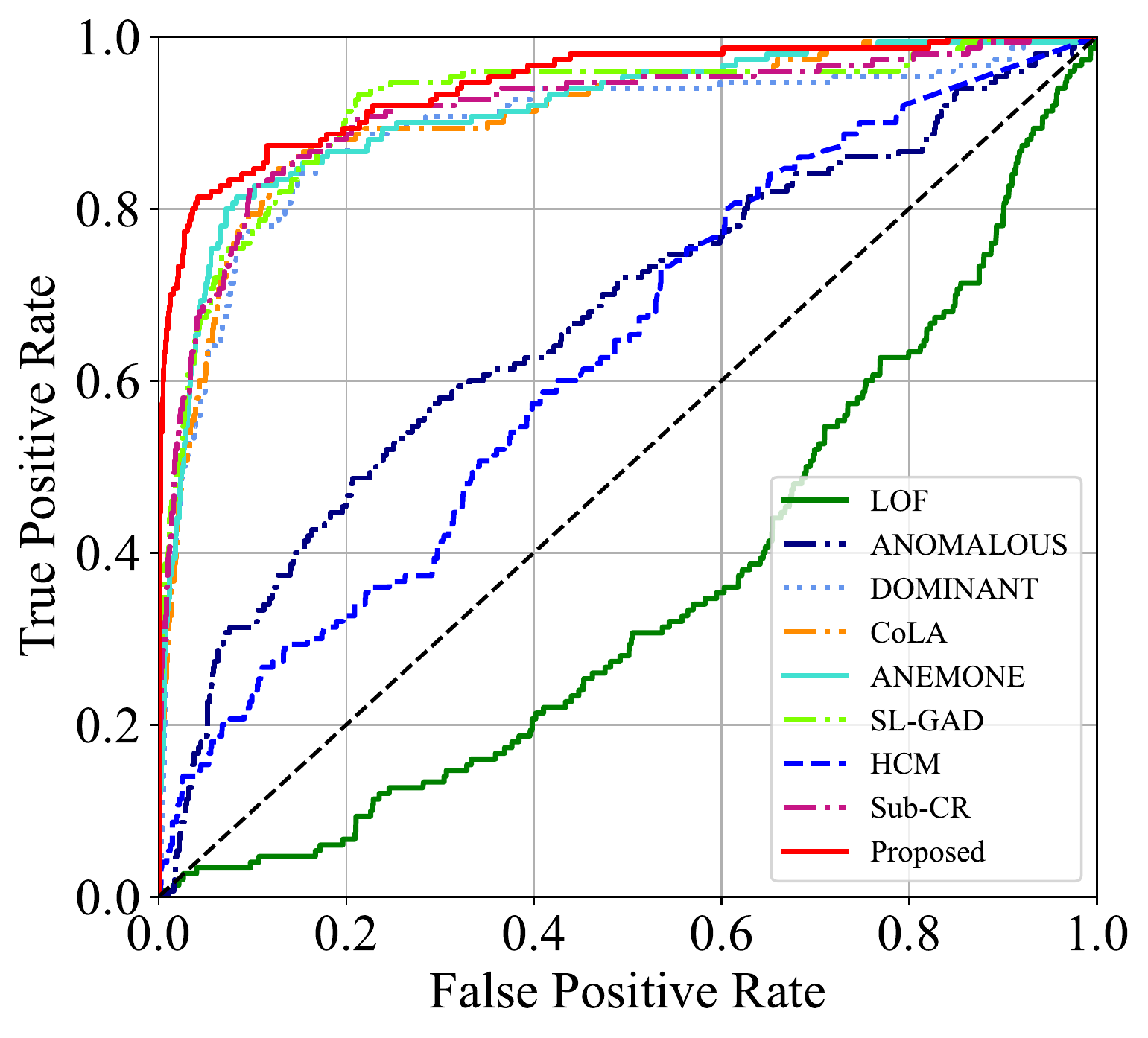}
}
\subfigure[ROC curve of CiteSeer.]{
\includegraphics[width=0.31\textwidth]{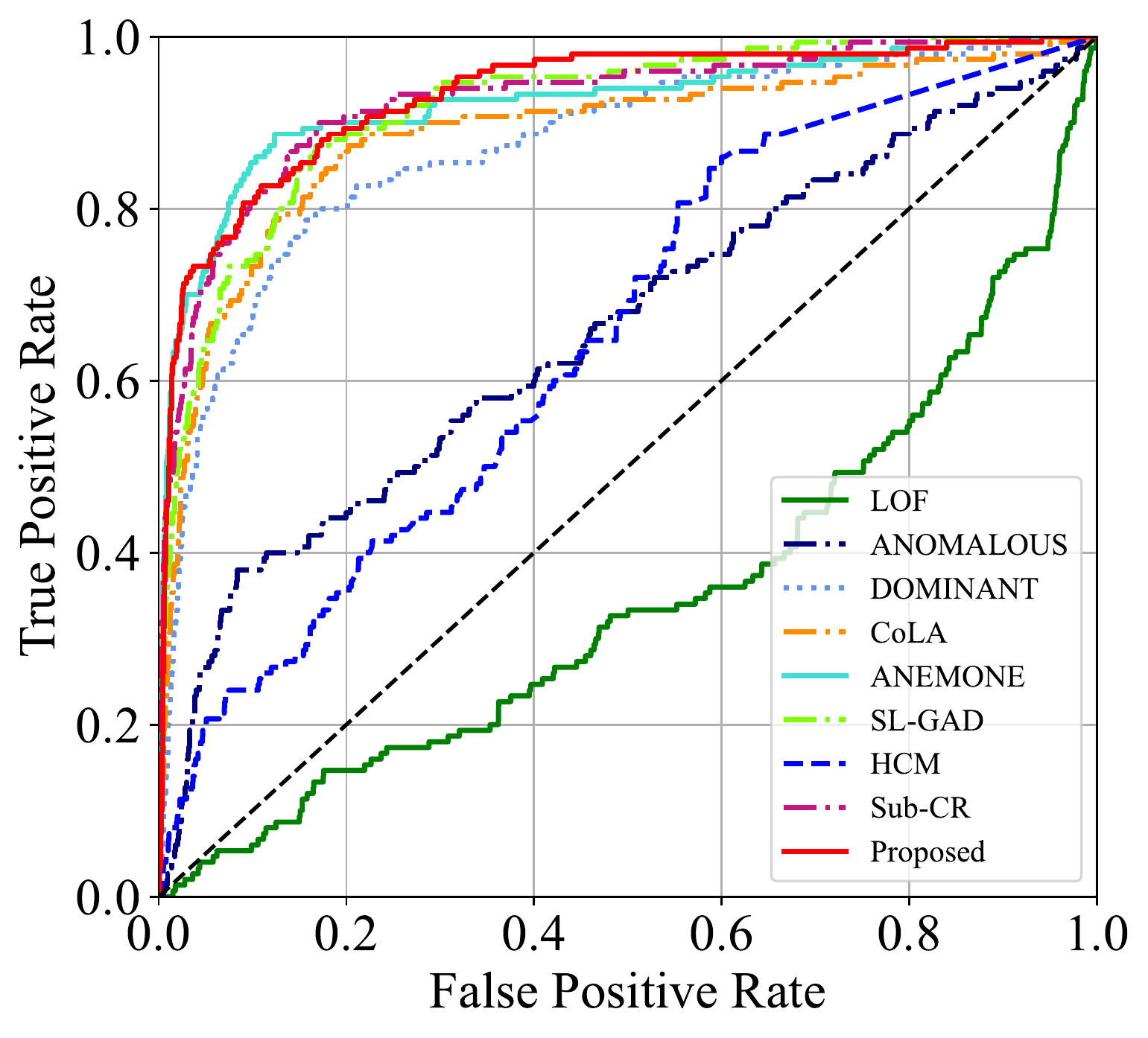}
}
\subfigure[ROC curve of DBLP.]{
\includegraphics[width=0.31\textwidth]{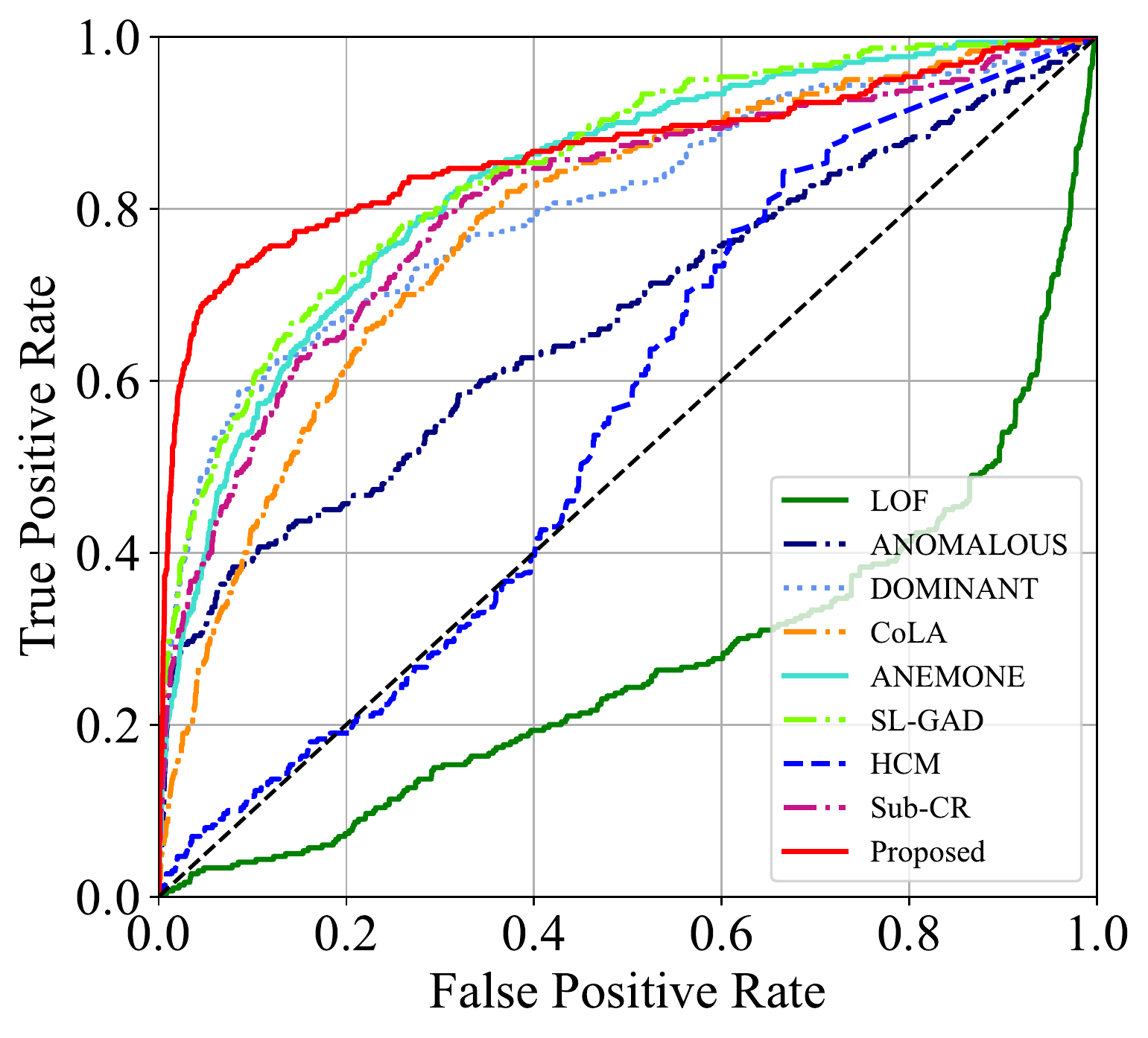}
}\\
\subfigure[ROC curve of Citation.]{
\includegraphics[width=0.31\textwidth]{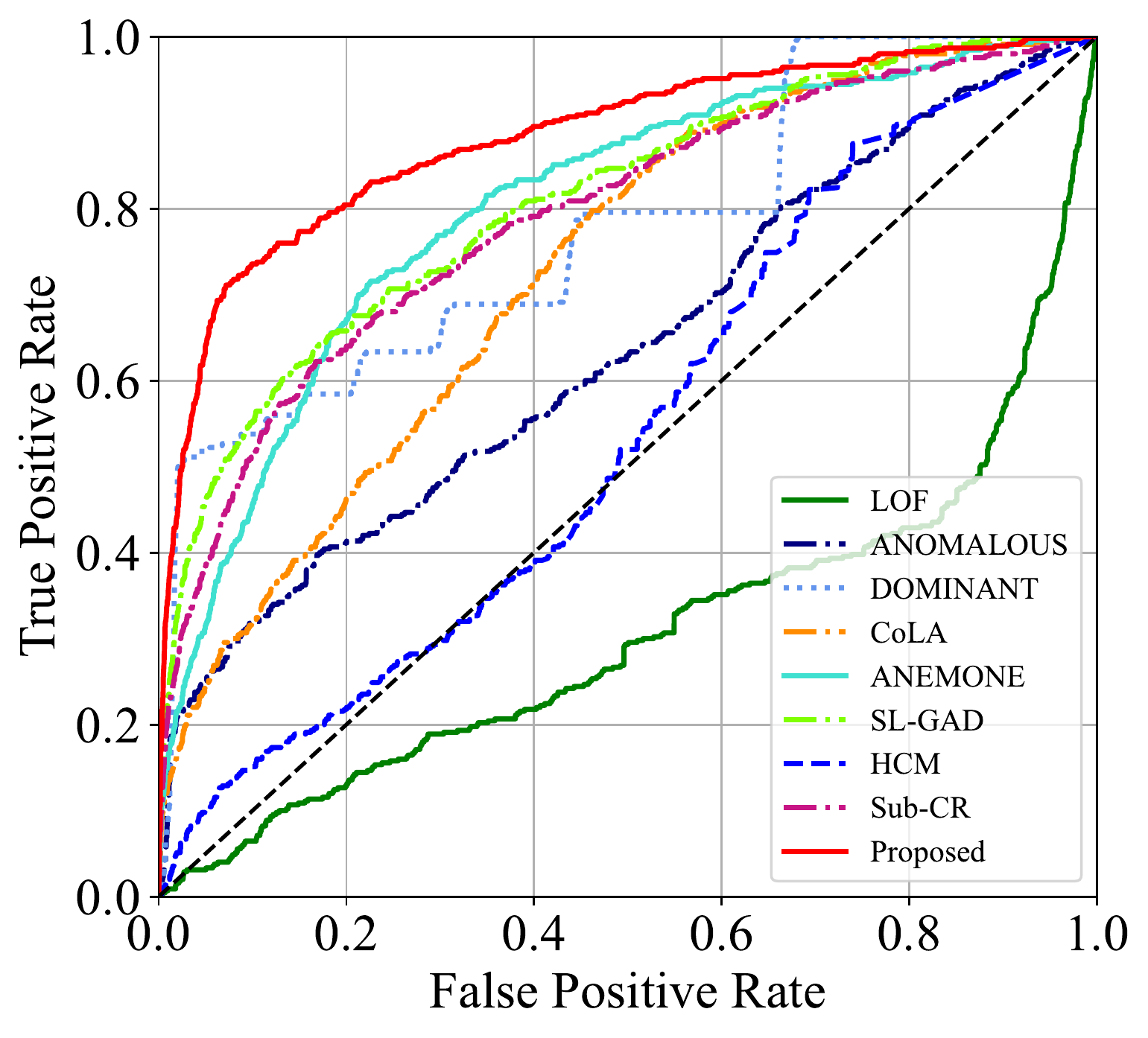}
}
\subfigure[ROC curve of ACM.]{
\includegraphics[width=0.31\textwidth]{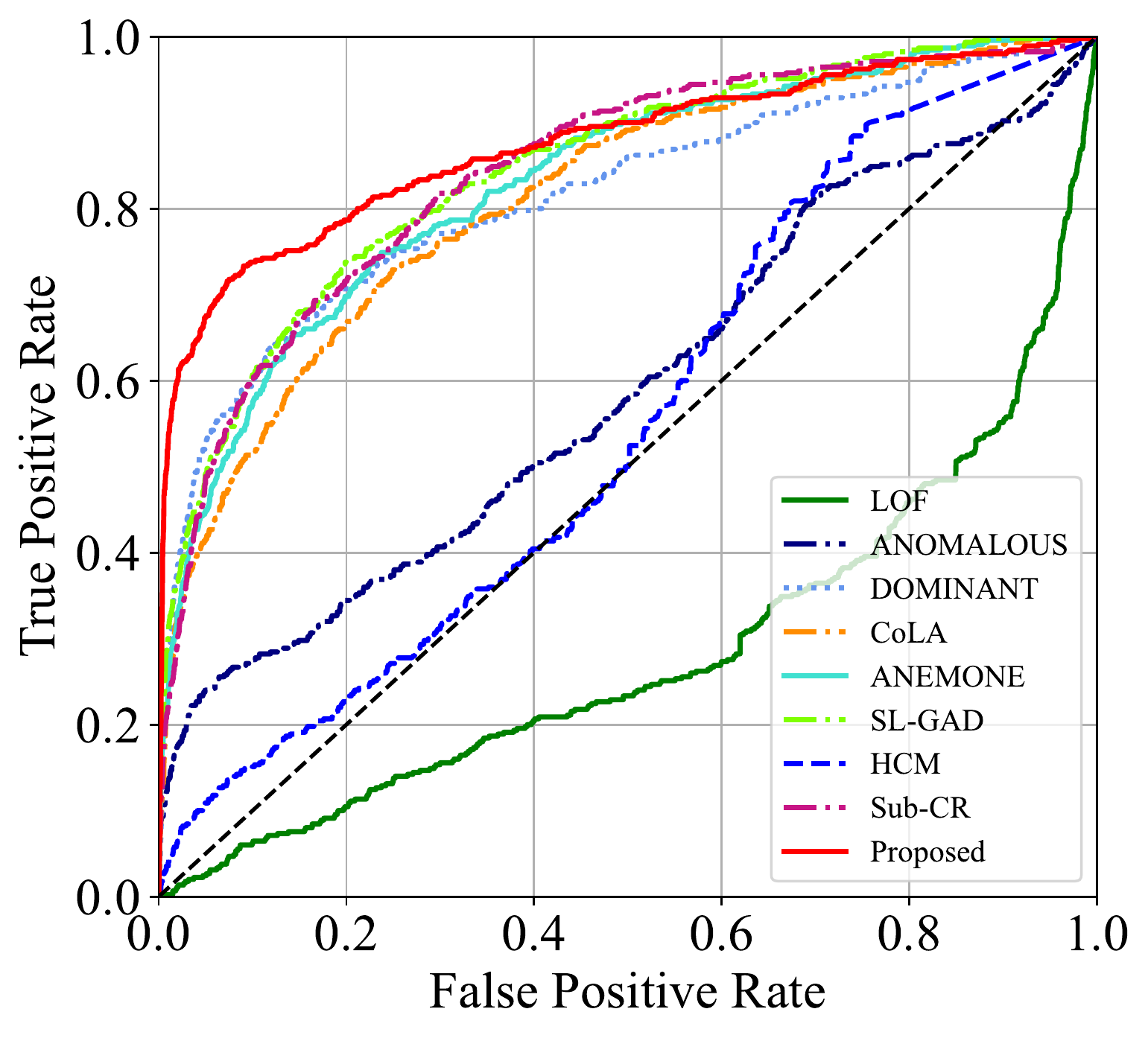}
}
\subfigure[ROC curve of PubMed.]{
\includegraphics[width=0.31\textwidth]{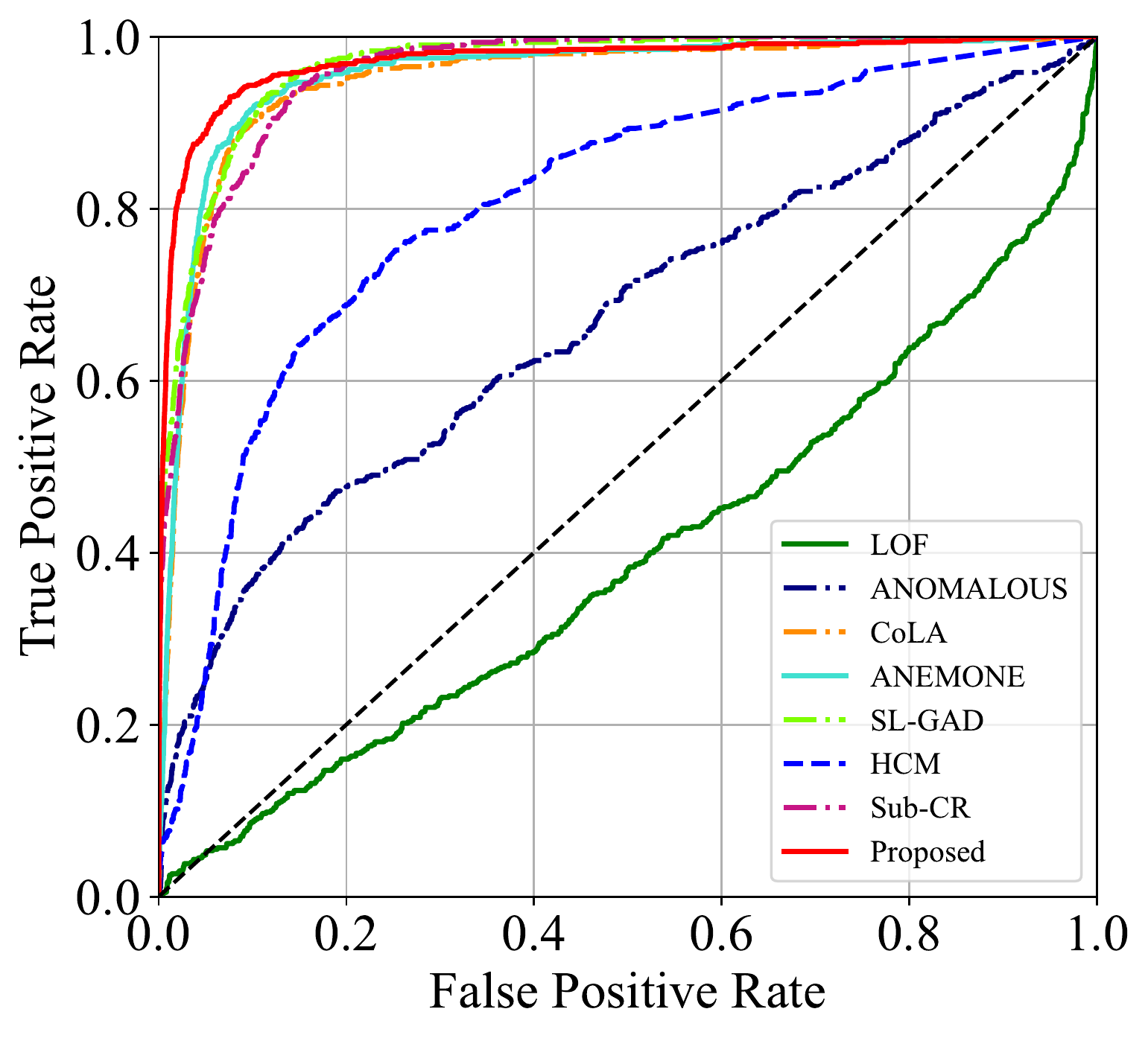}
}
% \vspace{5pt}
\caption{ROC curves comparison on six benchmark datasets. The area under the curve is larger, the anomaly detection performance is better. The black dotted lines are the ``random line'', indicating the performance under random guessing.}
\label{fig:AUC}
\end{figure*}

\begin{table*}[ht]
\vspace{-5pt}
\caption{Performance comparison for AUC and AUPRC. OOM indicates the issue of Out-Of-Memory happens during training. The bold and underlined values indicate the best and runner-up results, respectively.}
\resizebox{1.0\textwidth}{!}{
\begin{tabular}{ccccccccccccc}
\toprule
\multirow{2}{*}{\textbf{Methods}} & \multicolumn{2}{c}{\textbf{Cora}} & \multicolumn{2}{c}{\textbf{CiteSeer}} & \multicolumn{2}{c}{\textbf{DBLP}} & \multicolumn{2}{c}{\textbf{Citation}} & \multicolumn{2}{c}{\textbf{ACM}} & \multicolumn{2}{c}{\textbf{PubMed}}\\
\cmidrule(r){2-3}  \cmidrule(r){4-5} \cmidrule(r){6-7} \cmidrule(r){8-9} \cmidrule(r){10-11} \cmidrule(r){12-13}
& AUC & AUPRC & AUC & AUPRC & AUC & AUPRC & AUC & AUPRC & AUC & AUPRC & AUC & AUPRC \\
\midrule
LOF & 0.3538 & 0.0411 & 0.3484 & 0.0339 & 0.2694 & 0.0363 & 0.3059 & 0.0360 & 0.2843 & 0.0321 & 0.3934 & 0.0256 \\
ANOMALOUS & \textcolor{black}{0.6688} & \textcolor{black}{0.1181} & \textcolor{black}{0.6581} & \textcolor{black}{0.1084} & \textcolor{black}{0.6728} & \textcolor{black}{0.2444} & \textcolor{black}{0.6356} & \textcolor{black}{0.1413} & \textcolor{black}{0.5894} & \textcolor{black}{0.1752} & \textcolor{black}{0.6682} & \textcolor{black}{0.1096} \\
DOMINANT & 0.8929 & 0.4474 & 0.8718 & 0.3308 & 0.8034 & 0.3738 & 0.7748 & 0.2773 & 0.8152 & 0.3553 & OOM & OOM \\
% \textcolor{black}{ResGCN} & \textcolor{black}{0.9438} & \textcolor{black}{0.1181} & \textcolor{black}{0.9310} & \textcolor{black}{0.1084} & \textcolor{black}{0.8646} & \textcolor{black}{0.2444} & \textcolor{black}{0.8825} & \textcolor{black}{0.1413} & \textcolor{black}{0.8742} & \textcolor{black}{0.1752} & \textcolor{black}{0.9712} & \textcolor{black}{0.1096} \\
CoLA & 0.9065 & 0.5065 & 0.8863 & 0.4303 & 0.7824 & 0.2051 & 0.7296 & 0.1928 & 0.8127 & 0.3214 & 0.9493 & 0.4169 \\
ANEMONE & 0.9122 & 0.5320 & 0.9227 & \textbf{0.6281} & 0.8322 & 0.3271 & 0.8028 & 0.2578 & 0.8300 & 0.3539 & 0.9552 & 0.4274 \\
SL-GAD & \underline{0.9192} & \underline{0.6022} & 0.9177 & 0.5263 & \underline{0.8461} & \underline{0.4061} & \underline{0.8095} & \underline{0.3473} & \underline{0.8450} & \underline{0.3886} & \underline{0.9660} & \underline{0.6351} \\
HCM & 0.6276 & 0.1090 & 0.6502 & 0.0934 & 0.5572 & 0.0695 & 0.5414 & 0.0612 & 0.5507 & 0.0674 & 0.8065 & 0.1400 \\
Sub-CR & 0.9133 & 0.5922 & \underline{0.9248} & \underline{0.6083} & 0.8061 & 0.3336 & 0.7903 & 0.2968 & 0.8428 & 0.3469 & 0.9594 & 0.5854 \\
\midrule
\textbf{Proposed} & \textbf{0.9438} & \textbf{0.7750} & \textbf{0.9310} & 0.5500 & \textbf{0.8646} & \textbf{0.5483} & \textbf{0.8825} & \textbf{0.4924} & \textbf{0.8742} & \textbf{0.5632} & \textbf{0.9712} & \textbf{0.7287}\\ \bottomrule
\end{tabular}}
\label{table:AUCAUPRC}
\end{table*}

\textbf{AUC and AUPRC results.} As shown in Table \ref{table:AUCAUPRC}, we observe that \textcolor{black}{ARISE} has a stronger ability to distinguish anomalous nodes from normal nodes than other models. In particular, \textcolor{black}{ARISE} attains a remarkable AUC gain of \textbf{2.46\%}, \textbf{0.62\%}, \textbf{1.85\%}, \textbf{7.30\%}, \textbf{2.92\%} and \textbf{0.52\%} on Cora, CiteSeer, DBLP, Citation, ACM, and PubMed, respectively. At the same time, the AUPRC results reflect that \textcolor{black}{ARISE} makes a remarkable improvement in the positive sample detection compared with its competitors on most datasets, and it obtains about \textbf{17.46\%} AUPRC increment on the ACM dataset. These indicate that \textcolor{black}{ARISE} has a more comprehensive performance and can adapt to the different needs of various scenarios. It is worth noting that the well-design topology anomaly detection module brings about \textbf{2.19\%-15.29\%} AUC and \textbf{11.97\%-34.32\%} AUPRC gains when compared with the pioneering baseline CoLA.

Figure~\ref{fig:AUC} offers an intuitive illustration that the under-curve areas of the deep methods are significantly larger than most non-deep methods, and the contrastive learning-based methods work best among the deep methods. This indicates that deep methods, especially contrastive learning-based methods, can learn more effective embeddings of node attributes than non-deep methods for the ANAD task.

\textbf{Precision@K results.} For a more comprehensive evaluation, we also report Precision@K on the six datasets in Figure~\ref{fig:precision}. For each dataset, the odd columns are Precision@K results reflecting the proportion of true total anomalies, which contains both attribute and topology anomalies, in its top $k$ ranked nodes. The even columns are Precision@K measuring the proportion of veritable topology anomalies in its top $k$ anomaly scores. As suggested by results in Figure~\ref{fig:precision}, \textcolor{black}{ARISE} achieves best or comparable results in odd column subfigures. It is noticeable that \textcolor{black}{ARISE} brings tangible Precision@K improvement to the competitors except on Citation in even column subfigures. These phenomena verify the effectiveness of our novel substructure-based paradigm on topology anomaly detection. \textcolor{black}{According to the odd column subfigures, the proposed framework has stronger comprehensive detection performance than competitors.}

\textcolor{black}{
\subsubsection{Study on Topology Anomaly Injection} 
Currently, there are two major topology anomaly injection methods. The first one is making randomly selected nodes \textbf{fully connected}~\cite{ding2019interactive}. The other is to randomly \textbf{drop out} partial links between the selected nodes after the operation of the first injection method~\cite{liu2022bond}. Our paper adopts the first method in the experiments, which has been widely used in~\cite{ding2019deep, liu2021anomaly,jin2021anemone, zheng2021generative, zhang2022reconstruction}.}

\textcolor{black}{We perform comparative experiments to further investigate the generalization ability of our model on topology anomaly detection. To be specific, we follow the second injection method of topology anomalies, which drops out 10\% edges between the randomly selected nodes (the injection of attribute anomalies remains unchanged). \textbf{As shown in Table~\ref{table:new_inject}, our model still outperforms the competitors on all experimental datasets.} Besides, we observe that most methods suffer from performance degradation under the new injection condition. This means such an injection method would make it more difficult to discern topology anomalies from normal nodes. This will be one of the challenges for future work.}

\begin{table}[ht]
\centering
\caption{Performance comparison for AUC under different topology injection settings (drop out 10\% edges between selected nodes). OOM indicates the issue of Out-Of-Memory happens during training. The bold and underlined values indicate the best and runner-up results, respectively.}
\resizebox{0.48\textwidth}{!}{
\begin{tabular}{ccccccc}
\toprule
\textbf{Methods}  & \textbf{Cora}   & \textbf{CiteSeer} & \textbf{DBLP}   & \textbf{Citation} & \textbf{ACM}    & \textbf{PubMed} \\ \midrule
LOF               & 0.3527          & 0.3584            & 0.2905          & 0.2860            & 0.3038          & 0.3809          \\
ANOMALOUS         & 0.6781          & 0.6644            & 0.6752          & 0.6254            & 0.6375          & 0.6587          \\
DOMINANT          & 0.8958          & 0.8923            & 0.7916          & 0.7713            & 0.7894          & OOM             \\
CoLA              & 0.8991          & 0.8984            & 0.7861          & 0.7654            & 0.7922          & 0.9444          \\
ANEMONE           & \underline{0.9267}    & 0.9095            & 0.8170          & 0.7912            & 0.8009          & 0.9495          \\
SL-GAD            & 0.9072          & 0.9008            & 0.8142          & 0.8054            & 0.8247          & 0.9608          \\
HCM               & 0.6471          & 0.6432            & 0.5963          & 0.5743            & 0.5870          & 0.7949          \\
Sub-CR            & 0.9059          & \underline{0.9221}      & \underline{0.8204}    & \underline{0.8153}      & \underline{0.8336}    & \underline{0.9625}    \\ \midrule
\textbf{Proposed} & \textbf{0.9522} & \textbf{0.9297}   & \textbf{0.8374} & \textbf{0.8240}   & \textbf{0.8390} & \textbf{0.9627} \\ \bottomrule
\end{tabular}}
\label{table:new_inject}
\end{table}

\begin{figure}[ht]
\centering
\includegraphics[width = 0.33\textwidth]{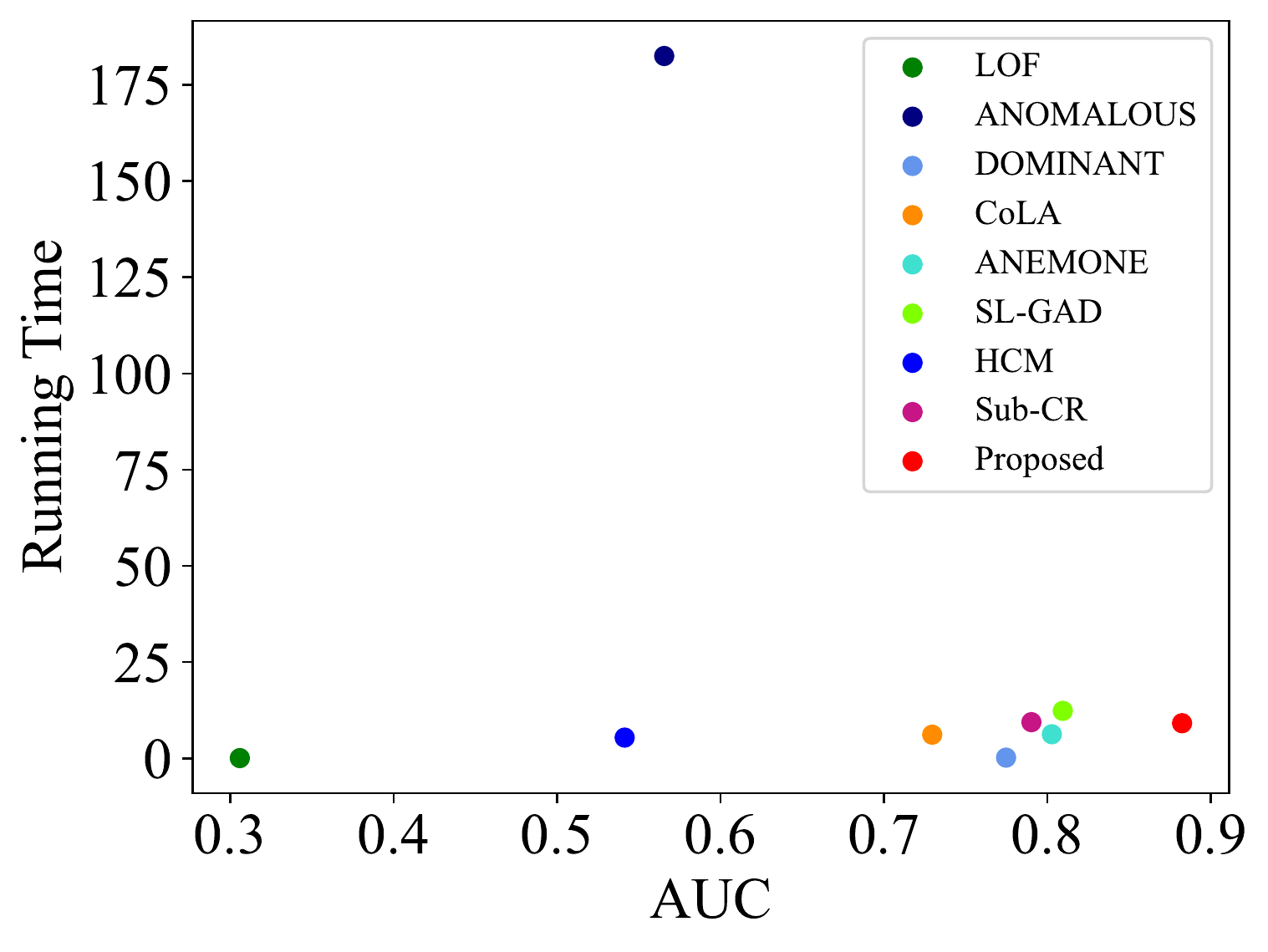}
\caption{\textcolor{black}{The relationships between AUC result and running time for all models on Citation. The running time results (second) are calculated by averaging the running time for each epoch out of 100 epochs.}}
\label{fig:time}
\vspace{-10pt}
\end{figure}

\textcolor{black}{
\subsubsection{Analysis of Time Efficiency} We further concretely investigate the time efficiencies of all methods. In practice, we compute the average running time (second) for each epoch out of 100 epochs. Figure~\ref{fig:time} shows the relationships between performance and running time (second) for all methods on Citation. Most methods spend near time, while ANOMALOUS runs significantly longer. And our model takes comparable time and achieves the best performance.}

\textcolor{black}{\subsection{Ablation Study}}
\textcolor{black}{In this subsection, we investigate the effects of the node representation acquisition strategies, the substructure detection method in the topology anomaly module, the topology anomaly score calculation strategy, and the score-fusion strategy on the model performance.}

\subsubsection{Analysis of Node Representations Acquisition}
In the topology anomaly module, the inner-node embedding similarities are highly correlated with the anomalous degree of the substructure. Hence, how to enjoy better node representations becomes a vital problem. We compare the performance of using raw node attributes with node embeddings from the contrastive network. Table~\ref{table:embedding} shows that the contrastive network learns more effective node representations for topology anomaly detection, further enhancing the ANAD performance.

\begin{center}
% \vspace{-10pt}
\begin{table}[ht]
\caption{Ablation analysis for the method of obtaining node representations w.r.t. AUC values. The bold values indicate the best results.}
\resizebox{0.48\textwidth}{!}{
\begin{tabular}{ccccccc}
\toprule
& \textbf{Cora} & \textbf{CiteSeer} & \textbf{DBLP} & \textbf{Citation} & \textbf{ACM} & \textbf{PubMed} \\
\midrule
Raw Node Attributes & 0.9270 & 0.8992 & 0.8623 & 0.8721 & 0.8682 & 0.9663 \\
\textbf{ARISE(Contrastive Network)} &\textbf{0.9437} & \textbf{0.9310} & \textbf{0.8646} & \textbf{0.8825} & \textbf{0.8742} & \textbf{0.9712}  \\ 
\bottomrule 
\end{tabular}}
\label{table:embedding}
\vspace{-25pt}
\end{table}
\end{center}

\textcolor{black}{
\subsubsection{Analysis of Substructure Detection Method} In the meantime, we perform ablation study on the substructure detection method in the topology anomaly module with other seven classical community search algorithms. \textbf{EHPPG}~\cite{kernighan1970efficient} divides the graph into communities by iteratively swapping node pairs to reduce the edge cut between them. \textbf{Greedy Modularity}~\cite{clauset2004finding} modifies the community partition of nodes by a greedy pattern until it reaches the max modularity of the graph. \textbf{Label Propagation}~\cite{cordasco2010community} is a semi-synchronous label propagation method. \textbf{DANMF}~\cite{ye2018deep} detects communities in the graph by combining nonnegative matrix factorization and deep auto-encoder. \textbf{PercoMCV}~\cite{kasoro2019percomcv} first performs the clique percolation algorithm, then computes eigenvector centrality. \textbf{EdMot}~\cite{li2019edmot} designs an edge enhancement approach for motif-aware community detection. \textbf{DNTM}~\cite{jaiswal2021detecting} works based on a distributed neighbourhood threshold method. As shown in Table~\ref{table:community}, the adopted \textbf{$k$-core}~\cite{batagelj2003m} based method outperforms the other community search methods, which means it can more effectively capture suspicious substructures and work better with the framework.}

\begin{center}
\begin{table}[ht]
\centering
\caption{Ablation analysis for the substructure detection method in the topology anomaly detection module w.r.t. AUC values. The bold values indicate the best results.
% , and the bold values indicate comparable results.
}
\resizebox{0.48\textwidth}{!}{
\begin{tabular}{ccccccc}
\toprule
& \textbf{Cora} & \textbf{CiteSeer} & \textbf{DBLP} & \textbf{Citation} & \textbf{ACM} & \textbf{PubMed} \\
\midrule
EHPPG (1970) &0.9032 &0.8995 &0.8235 &0.8407 &0.8333 &0.9553\\
Greedy Modularity (2004) &0.9212 &0.9038 &0.8274 &0.8515 &0.8362 &0.9551\\
Label Propagation (2010) &0.9138 &0.9095 &0.8245 &0.8298 &0.8520 &0.9553\\
DANMF (2018) &0.9145 &0.8741 &0.8202 &0.8468 &0.8363 &0.9554\\
PercoMCV (2019) &0.9124 &0.9171 &0.8493 &0.8577 &0.8643 &0.9556\\
EdMot (2019) &0.9244 &0.8790 &0.8222 &0.8262 &0.8373 &0.9519\\    
DNTM (2021) &0.9164 &0.9009 &0.8236 &0.8295 &0.8315 &0.9541\\
% girvan newman &0.9320 &0.9037 &0.8488 &0.8420 &0.8600 &0.9688\\
\midrule
\textbf{ARISE($k$-core)} (2003) &\textbf{0.9438} &\textbf{0.9310} &\textbf{0.8646} &\textbf{0.8825} &\textbf{0.8742} &\textbf{0.9712}\\
\bottomrule 
\end{tabular}}
\label{table:community}
\vspace{-25pt}
\end{table}
\end{center}

\textcolor{black}{
\subsubsection{Analysis of Topology Anomaly Score Calculation} As for topology anomaly score calculation, the number of node pairs can also be used to measure the size of substructures in Eq.~\eqref{topo_score2}. To study the effectiveness of the averaging component in Eq.~\eqref{sim_comm}, we conduct additional experiments on all datasets. The comparison between the first and last row in Table~\ref{table:formula} indicates that the number of nodes works better in our scheme. In the meantime, compared with ``W/O Averaging'', our scheme achieves the best or comparable performance. The experiment results verify the effectiveness of our topology anomaly score calculation strategy.}

\begin{table}[ht]
\centering
\caption{Ablation analysis for the components of Eq.~\eqref{sim_comm} and Eq.~\eqref{topo_score2} w.r.t. AUC values. ``W/O Averaging'' means deleting the averaging term in Eq.~\eqref{sim_comm}. And ``Node Pair'' indicates replacing the number of nodes in the substructure with the number of node pairs. The bold and underlined values indicate the best and runner-up results, respectively.}
\resizebox{0.48\textwidth}{!}{
\begin{tabular}{ccccccc}
\toprule
& \textbf{Cora} & \textbf{CiteSeer} & \textbf{DBLP} & \textbf{Citation} & \textbf{ACM} & \textbf{PubMed} \\
\midrule
Node Pair &0.9320 &0.9037 &0.8488 &0.8420 &0.8600 &0.9682\\
W/O Averaging &0.9297 &\underline{0.9223} &0.8413 &0.8751 &\textbf{0.8893} &\underline{0.9699}\\    
Node Pair \& W/O Averaging &\underline{0.9395} &0.9282 &\underline{0.8640} &\underline{0.8797} &\underline{0.8835} &0.9691\\
\textbf{ARISE} &\textbf{0.9438} &\textbf{0.9310} &\textbf{0.8646} &\textbf{0.8825} &0.8742 &\textbf{0.9712}\\
\bottomrule 
\end{tabular}
}
\label{table:formula}
% \vspace{-10pt}
\end{table}

\subsubsection{Analysis of Fusion Strategy for Topology and Attribute Anomaly Scores}
After gathering important topology and attribute anomaly scores, we analyze the fusion strategies to get final anomaly scores. To boost the performance, we explore three different combination strategies: (1) Max: We choose the largest value to represent the node anomaly degree; (2) Sum: We sum the two scores directly; (3) Weight: The scores are summed by weight. As shown in Table \ref{table:fusion}, \textcolor{black}{ARISE} adopts the weight fusion strategy, which is not always the best but the most comprehensive and effective choice.

\begin{center}
\begin{table}[ht]
\caption{Ablation analysis for the fusion method of topology and attribute anomaly scores w.r.t. AUC values. The bold values indicate the best results.
}
\resizebox{0.48\textwidth}{!}{
\begin{tabular}{ccccccc}
\toprule
& \textbf{Cora} & \textbf{CiteSeer} & \textbf{DBLP} & \textbf{Citation} & \textbf{ACM} & \textbf{PubMed} \\
\midrule
Max &\textbf{0.9448} &0.9116 &0.8539 &0.8527 &0.8645 &0.9408\\
Sum &0.9280 &0.8922 &0.8356 &0.8421 &0.8391 &0.9266\\
\textbf{ARISE(Weight)} &0.9438 &\textbf{0.9310} &\textbf{0.8646} &\textbf{0.8825} &\textbf{0.8742} &\textbf{0.9712}\\
\bottomrule 
\end{tabular}}
\label{table:fusion}
% \vspace{-10pt}
\end{table}
\end{center}

\textcolor{black}{\subsection{Parameter Analysis}}
\textcolor{black}{In the following subsection, we analyze the influence of coefficient ${\alpha}$ and its related settings on the detection performance.} 

\subsubsection{Score-fusion Strategy}
\label{alpha_score_fusion}
\begin{figure}[ht]
    \centering
    \includegraphics[width = 0.33\textwidth]{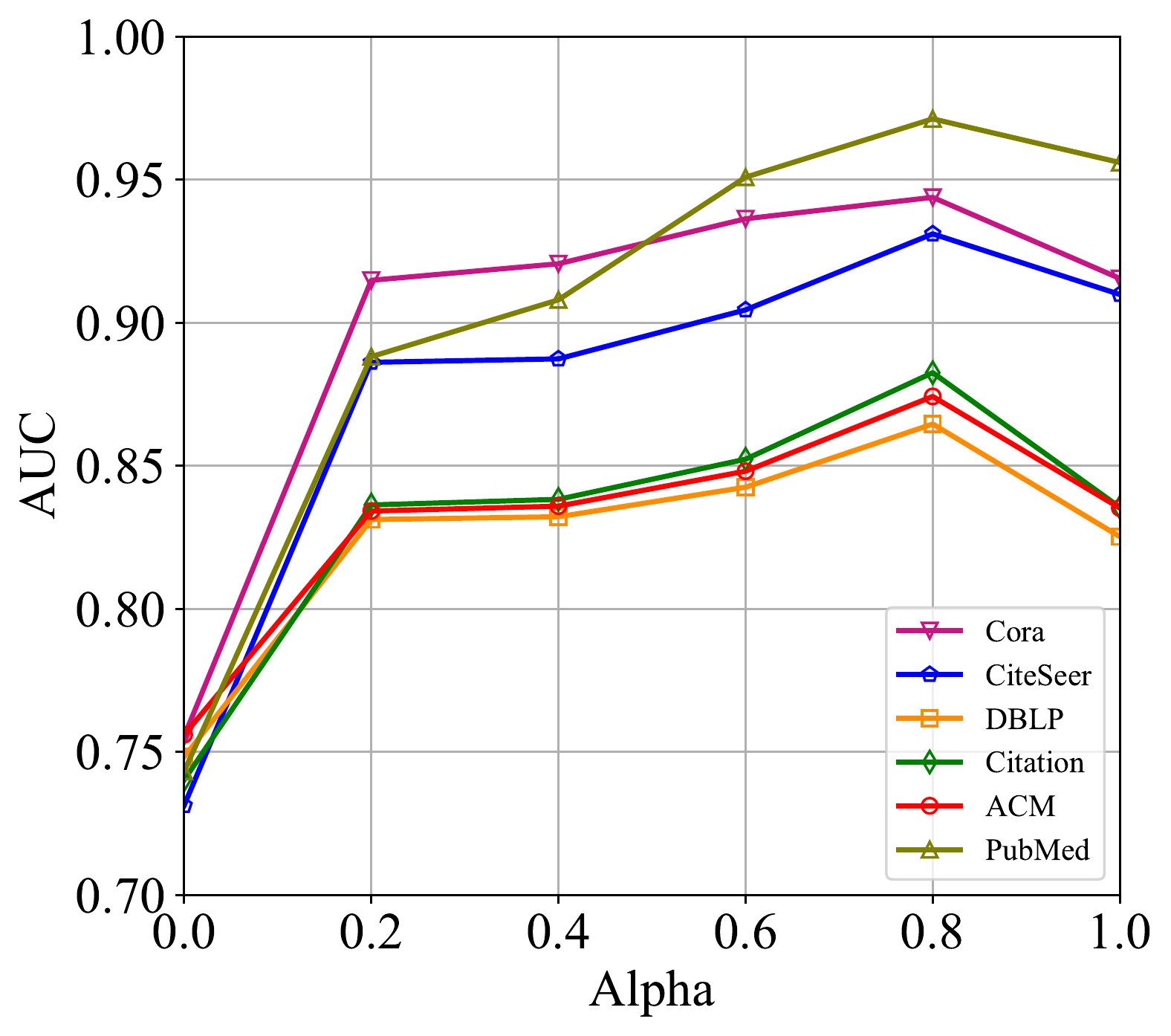}
    \caption{Trade-off parameter $\alpha$ w.r.t. AUC values.}
    \label{fig:Alpha}
\end{figure}

We first study the direct effect of ${\alpha}$ on the score-fusion strategy. As shown in Figure~\ref{fig:Alpha}, we vary the coefficient ${\alpha}$ from 0 to 1.0 and obtain its effect on the AUC value. Finally, we adopt ${\alpha}$ equal to 0.8 on all datasets. We note that the lines show an upward trend followed by a downward trend. The performance on ${\alpha = 0}$ or ${\alpha = 1}$ is significantly lower when ${\alpha}$ is between 0 and 1. This reflects that using the topology and attribute anomaly module concurrently greatly improves the detection ability of the model compared with only one.

\begin{figure}[!ht]
\centering
\subfigure[Citation ($\alpha=0.2$)]{    \includegraphics[width=1\linewidth]{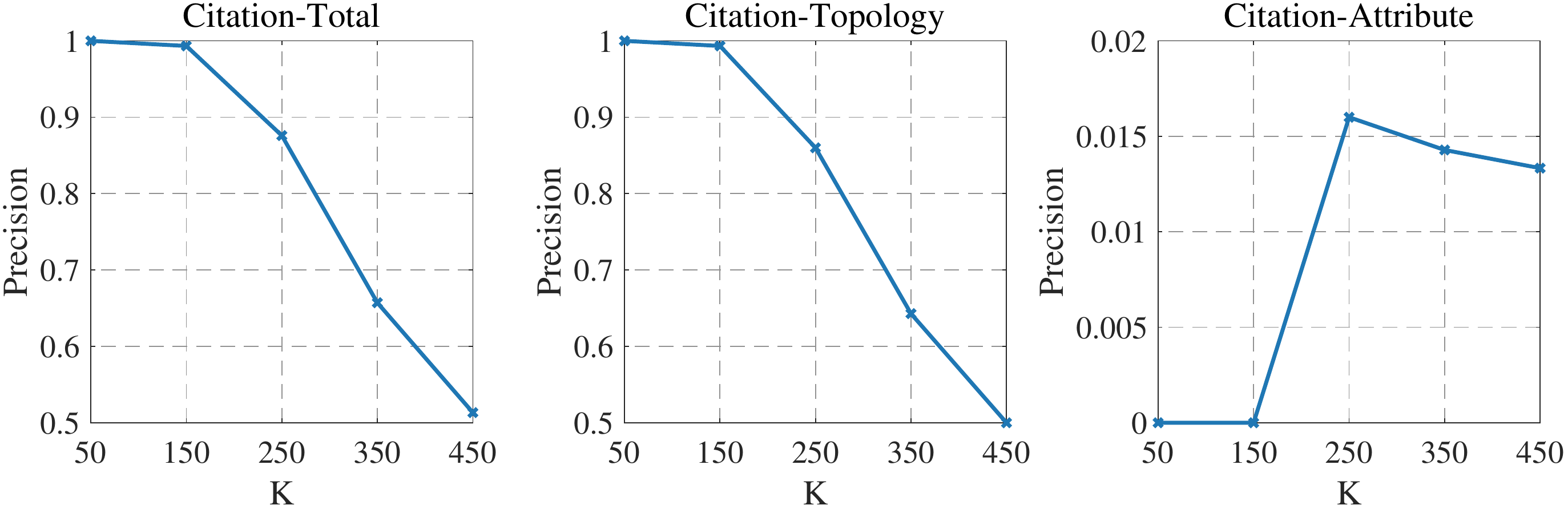}
}\\ 
\subfigure[Citation ($\alpha=0.8$)]{    \includegraphics[width=1\linewidth]{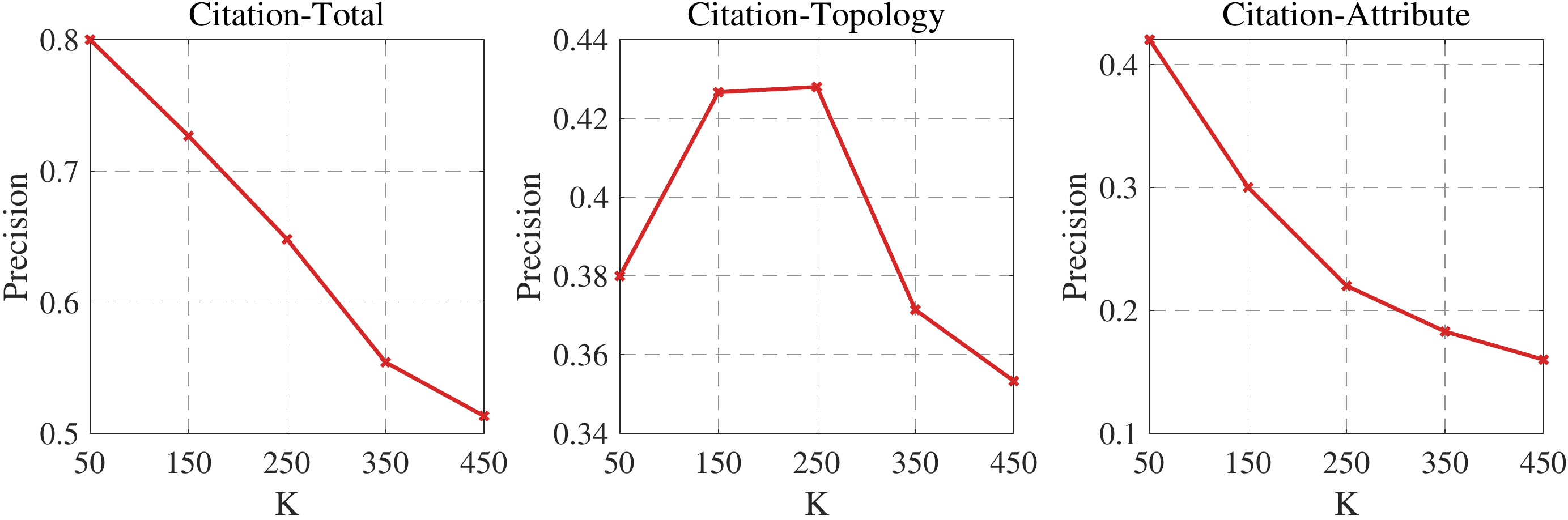}
}\\
\subfigure[PubMed ($\alpha=0.2$)]{    \includegraphics[width=1\linewidth]{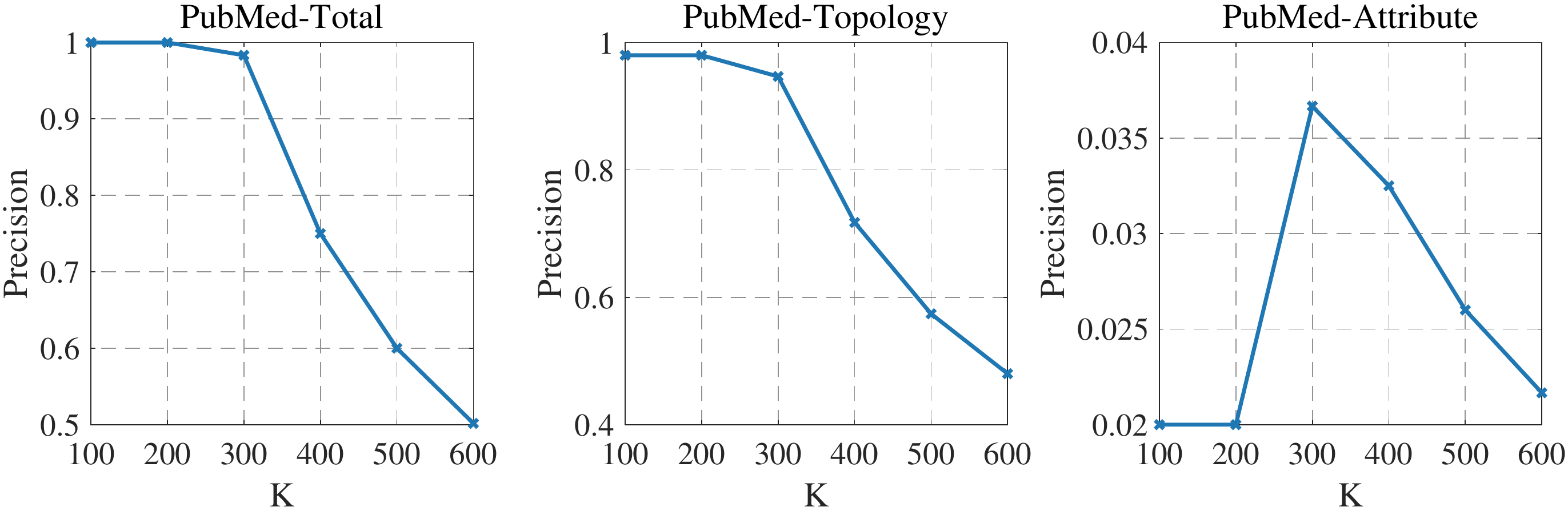}
}\\ 
\subfigure[PubMed ($\alpha=0.8$)]{    \includegraphics[width=1\linewidth]{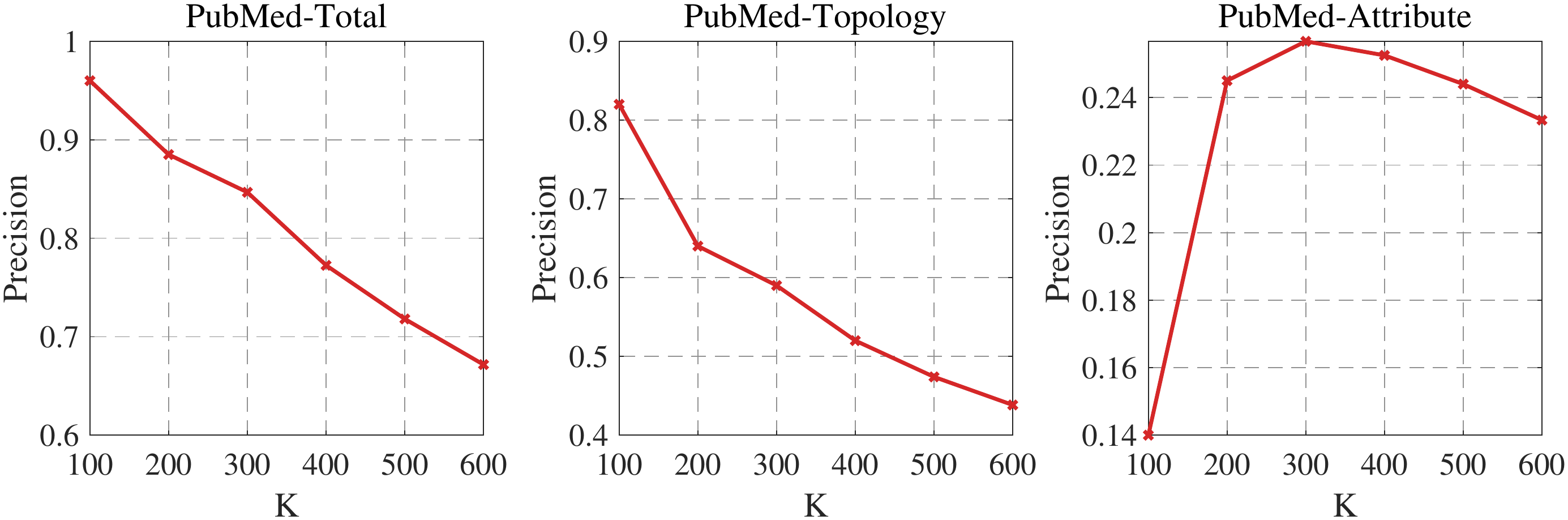}
}    
\caption{The impact of trade-off parameter $\alpha$ w.r.t. Precision@K of total anomaly, topology anomaly, and attribute anomaly on Citation and PubMed.}
\label{fig:precision_new}
\end{figure}

\textcolor{black}{To further explore this phenomenon, we report the Precision@K results of topology anomaly and attribute anomaly when $\alpha=0.2$ and $\alpha=0.8$ on Citation and PubMed in Figure~\ref{fig:precision_new}. Table~\ref{table:two_alpha} shows the AUC comparison of our model on different datasets. When $\alpha=0.2$, ARISE relies more on topology anomaly scores than attribute ones. In this case, Figure~\ref{fig:precision_new} indicates that the performance of detecting topology anomalies is much better than detecting attribute anomalies on all datasets. The model has a weak ability to detect attribute anomalies.}

\textcolor{black}{However, we aim to effectively detect such two types of anomalies simultaneously. Apparently, the model cannot achieve the goal when $\alpha=0.2$. The comparison of AUC results in Table~\ref{table:two_alpha} also validates it. On the contrary, $\alpha=0.8$ can balance the effects of two types of anomaly scores and enhance the detection performance of attribute anomalies. Under these circumstances, our model would have a more comprehensive detection performance. Therefore, we employ such a value strategy for $\alpha$.}
% , i.e. the attribute anomaly scores have more weight.}
% It is worth noting that $\alpha=0$ and $\alpha=1$ wound make the model lose anomalous information from an anomaly detection module.

\begin{table}[!ht]
\centering
\caption{The impact of trade-off parameter $\alpha$ w.r.t. AUC results.}
    % \resizebox{1\textwidth}{!}{
\resizebox{0.48\textwidth}{!}{
\begin{tabular}{c|cccccc}
\toprule
$\alpha$& \textbf{Cora} & \textbf{CiteSeer} & \textbf{DBLP} & \textbf{Citation} & \textbf{ACM} & \textbf{PubMed} \\
\midrule
0.2 &0.9148 &0.8861 &0.8312 &0.8363 &0.8341 &0.8881\\
0.8 &0.9438 &0.9310 &0.8646 &0.8825 &0.8742 &0.9712\\
\bottomrule
\end{tabular}
}
\label{table:two_alpha}
\vspace{-10pt}
\end{table}

\begin{figure*}[ht]
% \vspace{-5pt}
\centering
\subfigure[Ratio = 9:1]{
\includegraphics[width=0.3\linewidth]{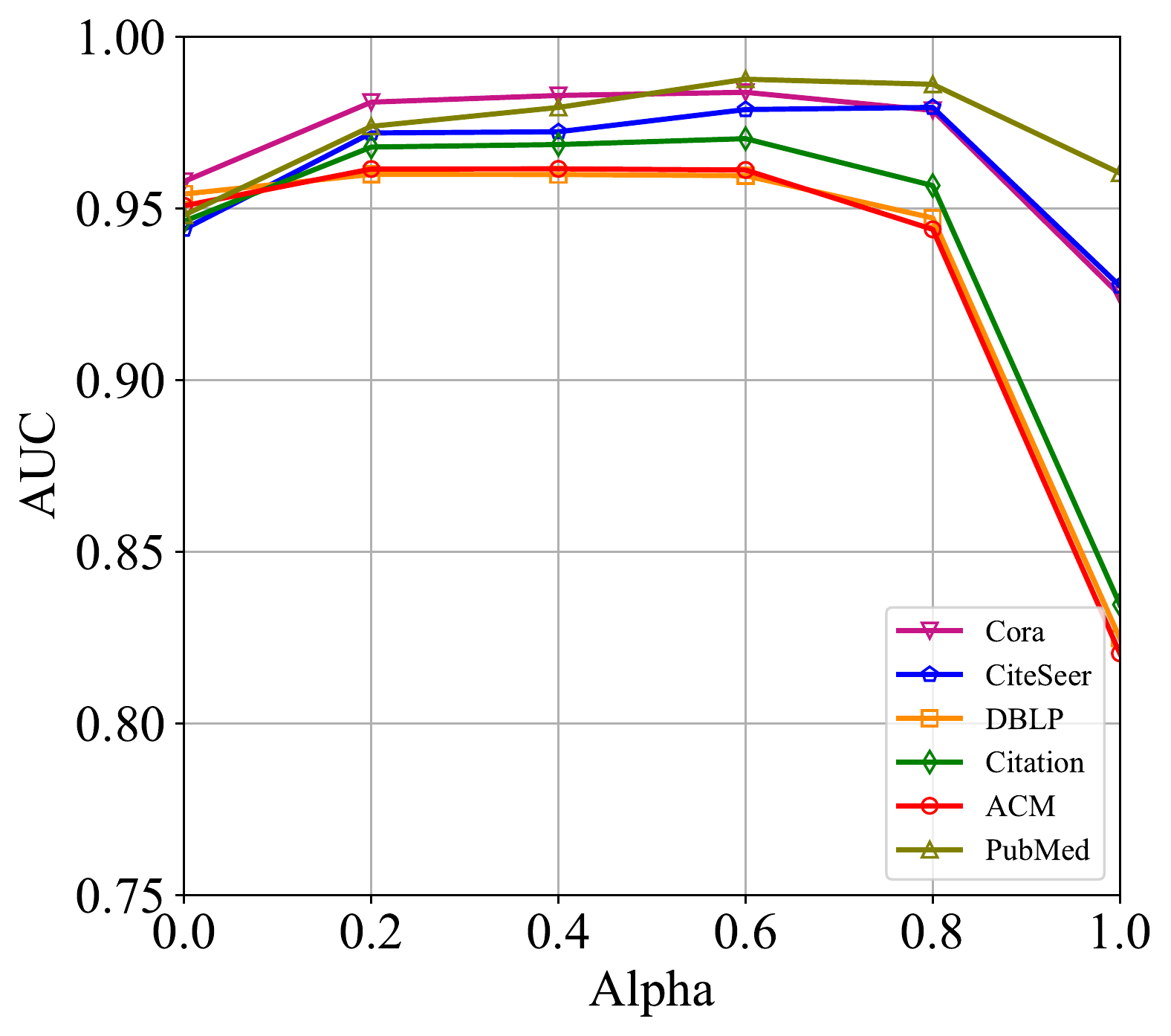}
\label{fig:alpha1}
}
\subfigure[Ratio = 7:3]{
\includegraphics[width=0.3\linewidth]{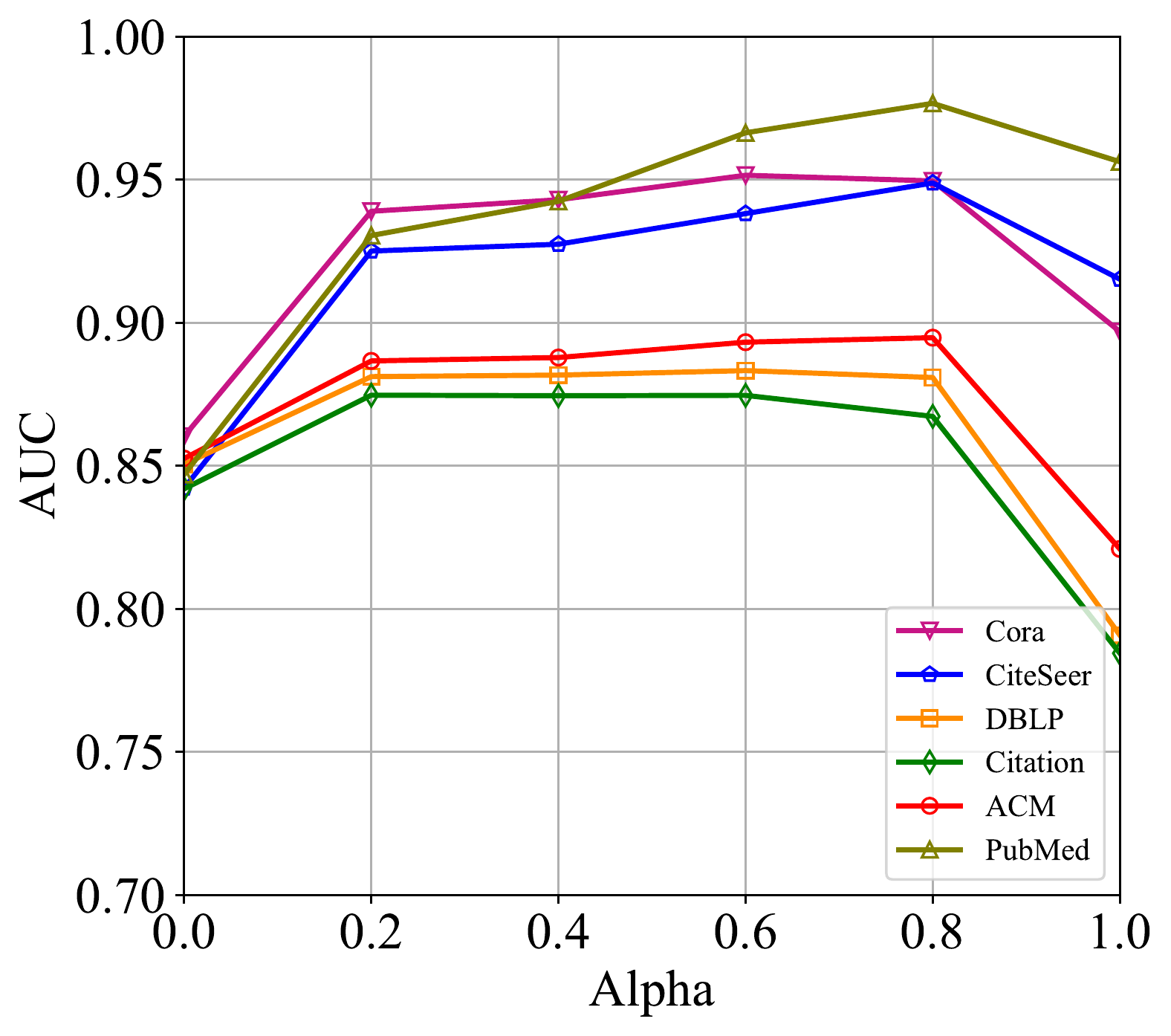}
\label{fig:alpha2}
}
\subfigure[Ratio = 1:1]{
\includegraphics[width=0.3\linewidth]{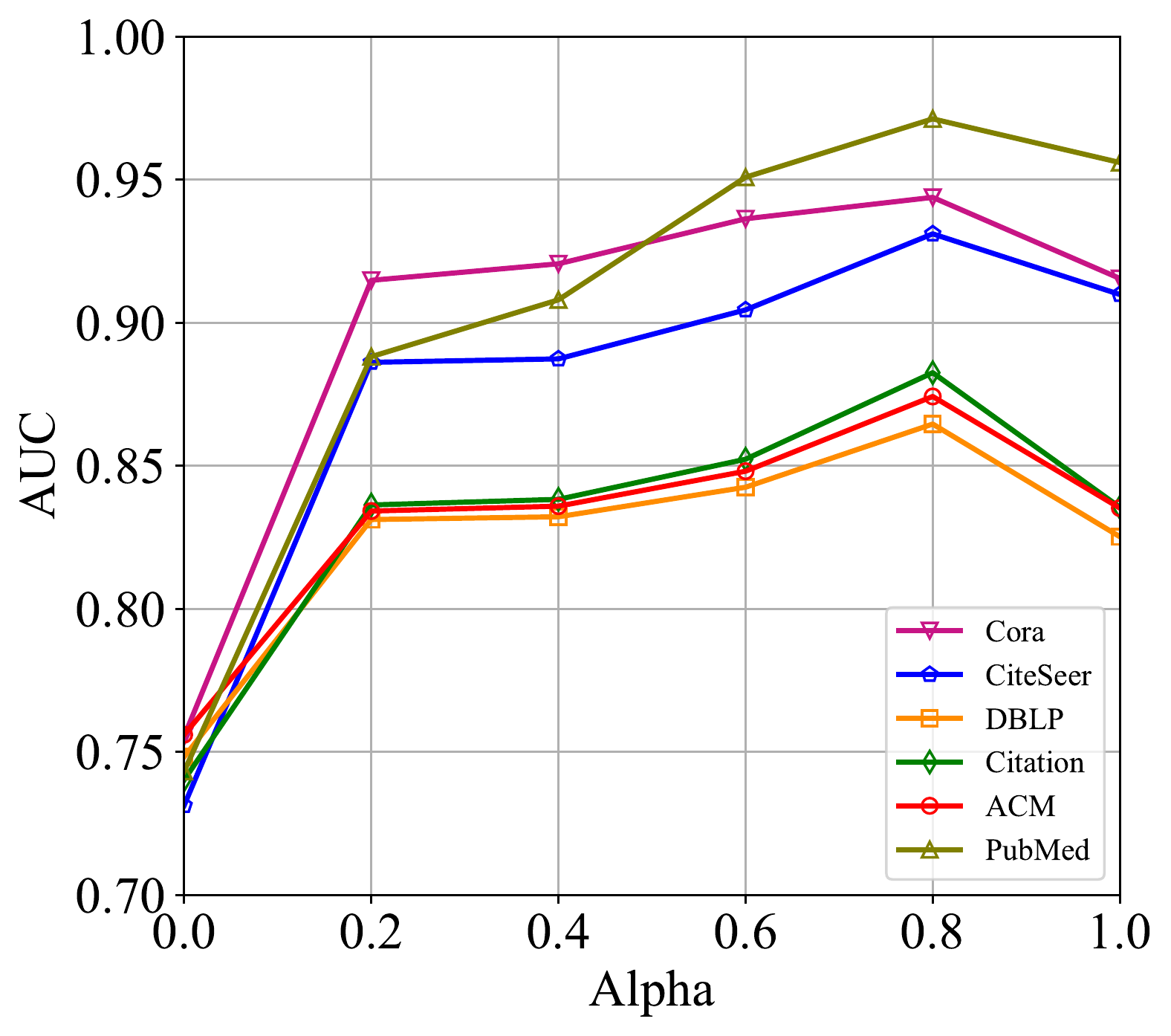}
\label{fig:alpha3}
}\\
\subfigure[Ratio = 3:7]{
\includegraphics[width=0.3\linewidth]{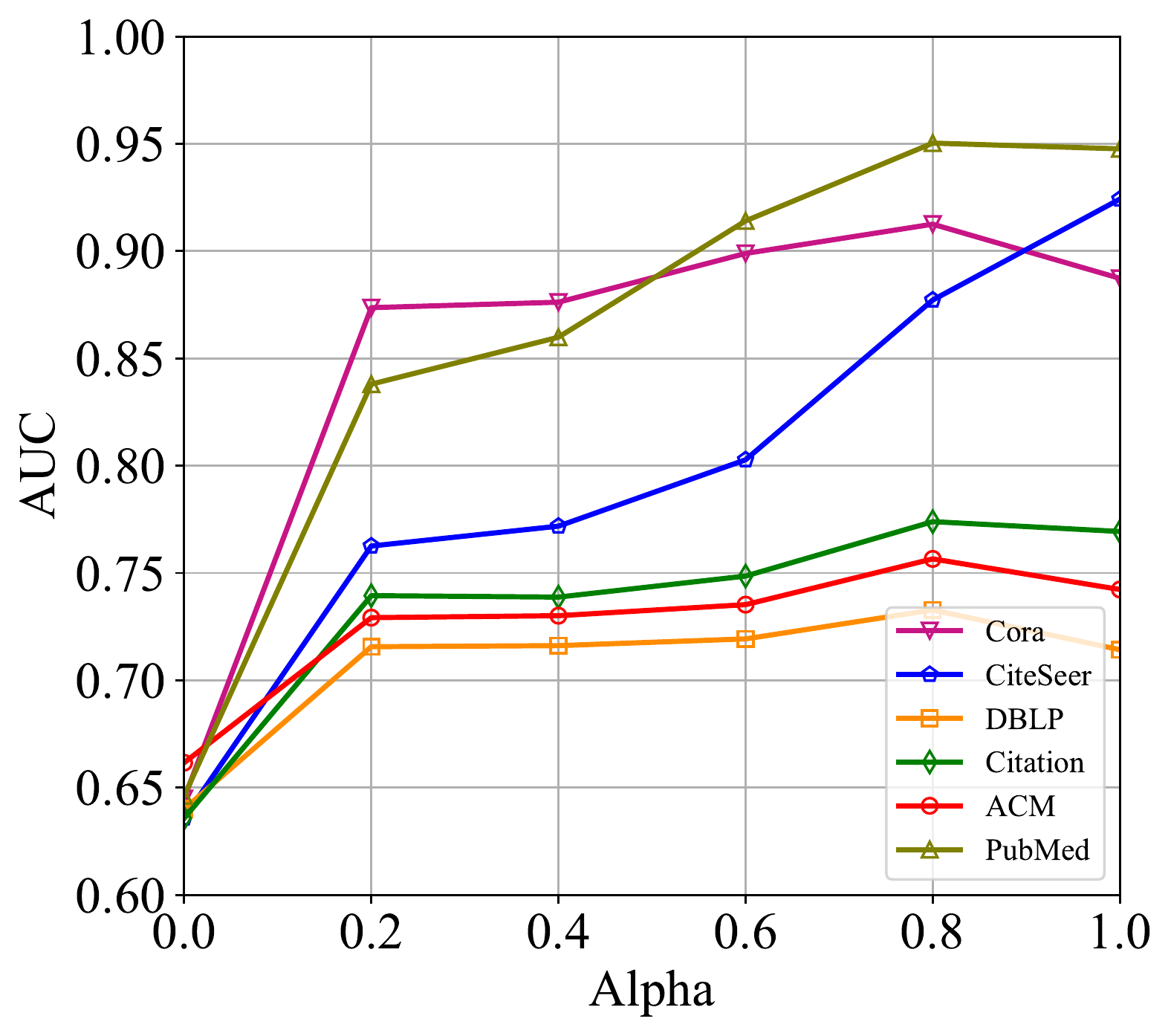}
\label{fig:alpha4}
}
\subfigure[Ratio = 1:9]{
\includegraphics[width=0.3\linewidth]{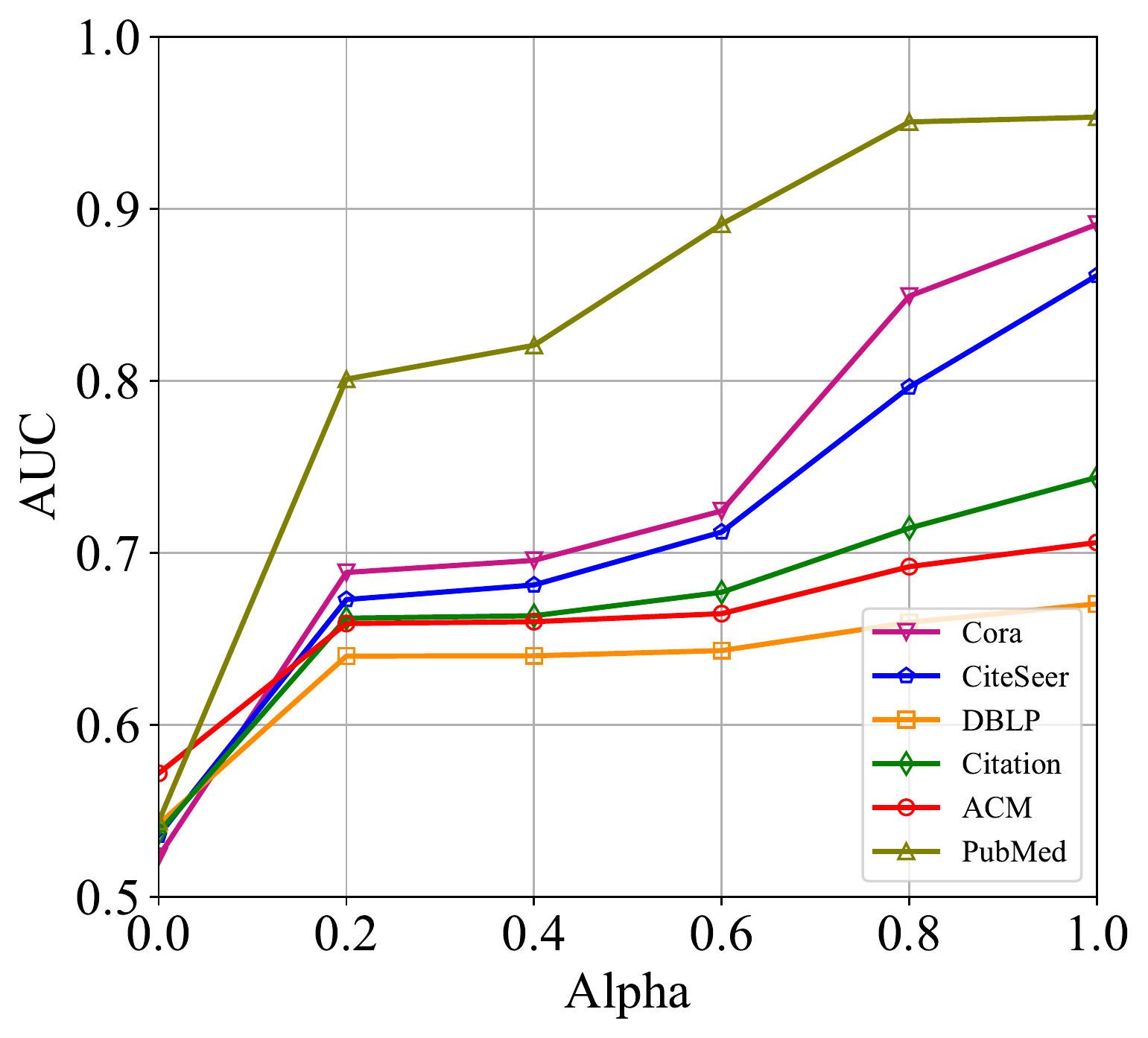}
\label{fig:alpha5}
}
\caption{The impact of trade-off parameter $\alpha$ w.r.t. AUC values under different ratios of topology and attribute anomalies.}
\label{fig:alpha}
% \vspace{-10pt}
\end{figure*}

\textcolor{black}{\subsubsection{Imbalanced Anomaly Injection} As mentioned in Subsection~\ref{injection}, we inject two types of anomalous nodes with the same number into the original graph. However, an imbalanced injection method might affect the selection of the trade-off parameter $\alpha$. We perform further experiments to study the relationship between the ratio of two types of nodes and $\alpha$. In practice, we design five different injection ratios of the topology anomalies and the attribute anomalies as shown in Table~\ref{table:TAAA}. The total number of anomalies remains unchanged in all datasets and we change the ratio according to the injection method in Subsection~\ref{injection}. Figure~\ref{fig:alpha} indicates the effect of the trade-off parameter $\alpha$ under different anomaly ratios on detection performance.}

\textcolor{black}{We can intuitively observe that the model achieves the best performance on most datasets when $\alpha=0.8$ in Subfigure~\ref{fig:alpha1}-\ref{fig:alpha4}. And all of the detection performances reach the peak when $\alpha=1.0$ in Subfigure~\ref{fig:alpha5}. As mentioned in Subsection~\ref{alpha_score_fusion}, $\alpha=0.8$ can provide a more comprehensive detection performance for such two types of anomalies in most cases. However, the number of attribute anomalies is nine times more than the number of topology ones under the last anomaly injection ratio. The model needs a stronger attribute anomaly detection performance under such an imbalanced injection. Hence, when $\alpha=1.0$, the model achieves the best performance.}

\begin{table*}[!ht]
\centering
\caption{The number of each type of anomalies injected into each dataset. ``TA'' represents the topology anomalies, ``AA'' indicates the attribute anomalies. ``Ratio'' is the ratio of topology anomalies and attribute anomalies.
% , and ``TA/AA'' is the result of the number of ``TA'' divided by the number of ``AA''.
}
% \resizebox{0.5\textwidth}{!}{
\begin{tabular}{c|cccccccccccc}
\toprule
\multirow{2}{*}{Ratio} & \multicolumn{2}{c}{\textbf{Cora}} & \multicolumn{2}{c}{\textbf{CiteSeer}} & \multicolumn{2}{c}{\textbf{DBLP}} & \multicolumn{2}{c}{\textbf{Citation}} & \multicolumn{2}{c}{\textbf{ACM}} & \multicolumn{2}{c}{\textbf{PubMed}}\\
\cmidrule(r){2-3}  \cmidrule(r){4-5} \cmidrule(r){6-7} \cmidrule(r){8-9} \cmidrule(r){10-11} \cmidrule(r){12-13}
& TA & AA & TA & AA & TA & AA & TA & AA & TA & AA & TA & AA \\
\midrule
9:1 & 135 & 15 & 135 & 15 & 270 & 30 & 405 & 45 & 405 & 45 & 540 & 60\\
7:3 & 105 & 45 & 105 & 45 & 210 & 90 & 315 & 135 & 315 & 135 & 420 & 180 \\
1:1 & 75 & 75 & 75 & 75 & 150 & 150 & 225 & 225 & 225 & 225 & 300 & 300 \\
3:7 & 45 & 105 & 45 & 105 & 90 & 210 & 135 & 315 & 135 & 315 & 180 & 420 \\
1:9 & 15 & 135 & 15 & 135 & 30 & 270 & 45 & 405 & 45 & 405 & 60 & 540 \\
\bottomrule
\end{tabular}
% }
\label{table:TAAA}
\vspace{-5pt}
\end{table*}

\section{Conclusion} 
In this paper, we propose a novel graph anomaly detection framework on attributed networks via substructure awareness termed ARISE. Different from existing works, we design a new topology anomaly detection principle from the perspective of network substructure. Specifically, we use a density-based algorithm to detect high-density substructures as the region proposed. \textcolor{black}{Then, the average node-pair similarity of nodes is measured to discern topology anomalies.} For this purpose, we introduce a graph contrastive learning method to obtain better embeddings of node attributes and to detect attribute anomalies. The experiments demonstrate that \textcolor{black}{ARISE} makes a great improvement on attributed networks anomaly detection. In future work, we continue to explore how to better aggregate the attribute and topology anomaly scores of nodes to avoid information loss.

\ifCLASSOPTIONcompsoc
  \section*{Acknowledgments}
\else
  \section*{Acknowledgment}
\fi

This work was supported by the National Key R\&D Program of China (project no. 2020AAA0107100) and the National Natural Science Foundation of China (project no. 62325604 and 62276271).

\ifCLASSOPTIONcaptionsoff
    \newpage
\fi

\bibliographystyle{IEEEtran}
\bibliography{acmart}

\vspace{-70pt}
\begin{IEEEbiography}[{\includegraphics[width=1in,height=1.25in,clip,keepaspectratio]{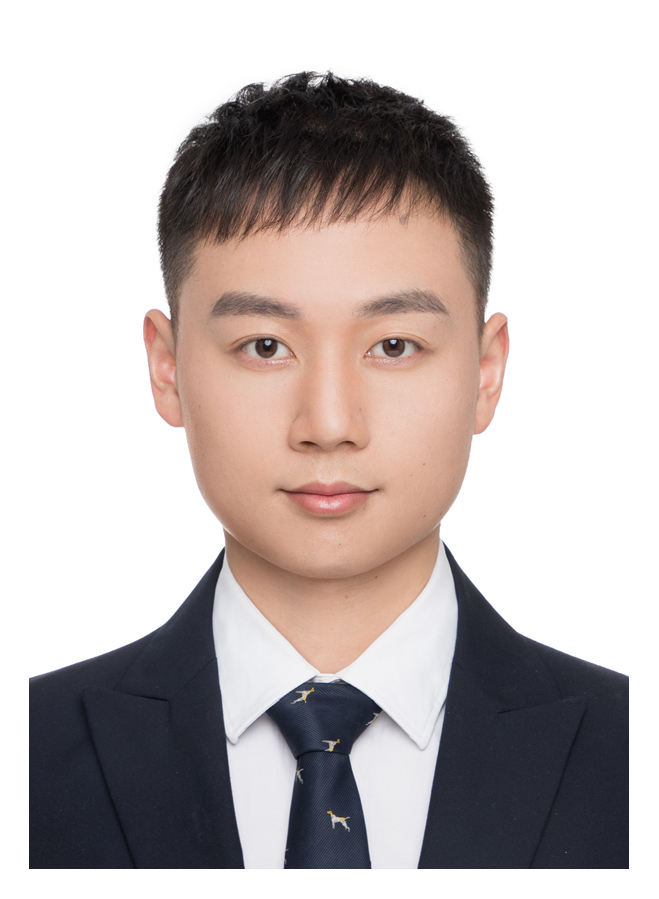}}]
{Jingcan Duan} is currently pursuing his master degree in the College of Computer Science and Technology, National University of Defense Technology (NUDT), Changsha, China.

His current research interests include graph representation learning, graph anomaly detection, and artificial intelligence (AI) for science.
\end{IEEEbiography}

\vspace{-40pt}
\begin{IEEEbiography}
[{\includegraphics[width=1in,height=1.25in,clip,keepaspectratio]{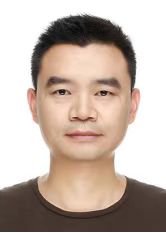}}]{Bin Xiao} received the B.S. and M.S. degrees in electrical engineering from Shanxi Normal University, Xi'an, China, in 2004 and 2007, respectively, and the Ph.D. degree in computer science from Xidian University, Xi'an, in 2012.

He is currently a Professor with the Chongqing University of Posts and Telecommunications, Chongqing, China. His research interests include image processing and pattern recognition.
% \vspace{-450pt}
\end{IEEEbiography}
\vspace{-40pt}

\begin{IEEEbiography}[{\includegraphics[width=1in,height=1.25in,clip,keepaspectratio]{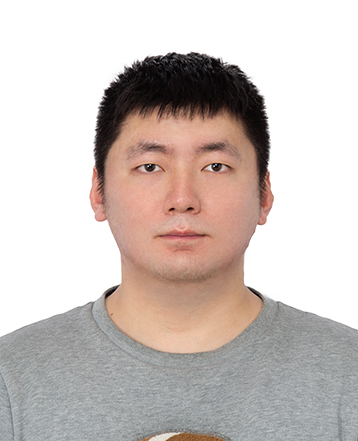}}]
{Siwei Wang} received the Ph.D. degree with the National University of Defense Technology (NUDT), Changsha, China, in 2023. 

He is currently an Assistant Research Professor with the Intelligent Game and Decision Laboratory, China. He has published several papers. His current research interests include kernel learning, unsupervised multiple-view learning, scalable clustering, and deep unsupervised learning.

Dr. Wang served as a PC Member/Reviewer in top journals and conferences, such as IEEE TRANSACTIONS ON KNOWLEDGE AND DATA ENGINEERING (TKDE), IEEE TRANSACTIONS ON NEURAL NETWORKS AND LEARNING SYSTEMS (TNNLS), IEEE TRANSACTIONS ON IMAGE PROCESSING (TIP), IEEE TRANSACTIONS ON CYBERNETICS (TCYB), IEEE TRANSACTIONS ON MULTIMEDIA (TMM), ICML, CVPR, ECCV, ICCV, AAAI, and IJCAI.
\end{IEEEbiography}
\vspace{-40pt}

\begin{IEEEbiography}[{\includegraphics[width=1in,height=1.25in,clip,keepaspectratio]{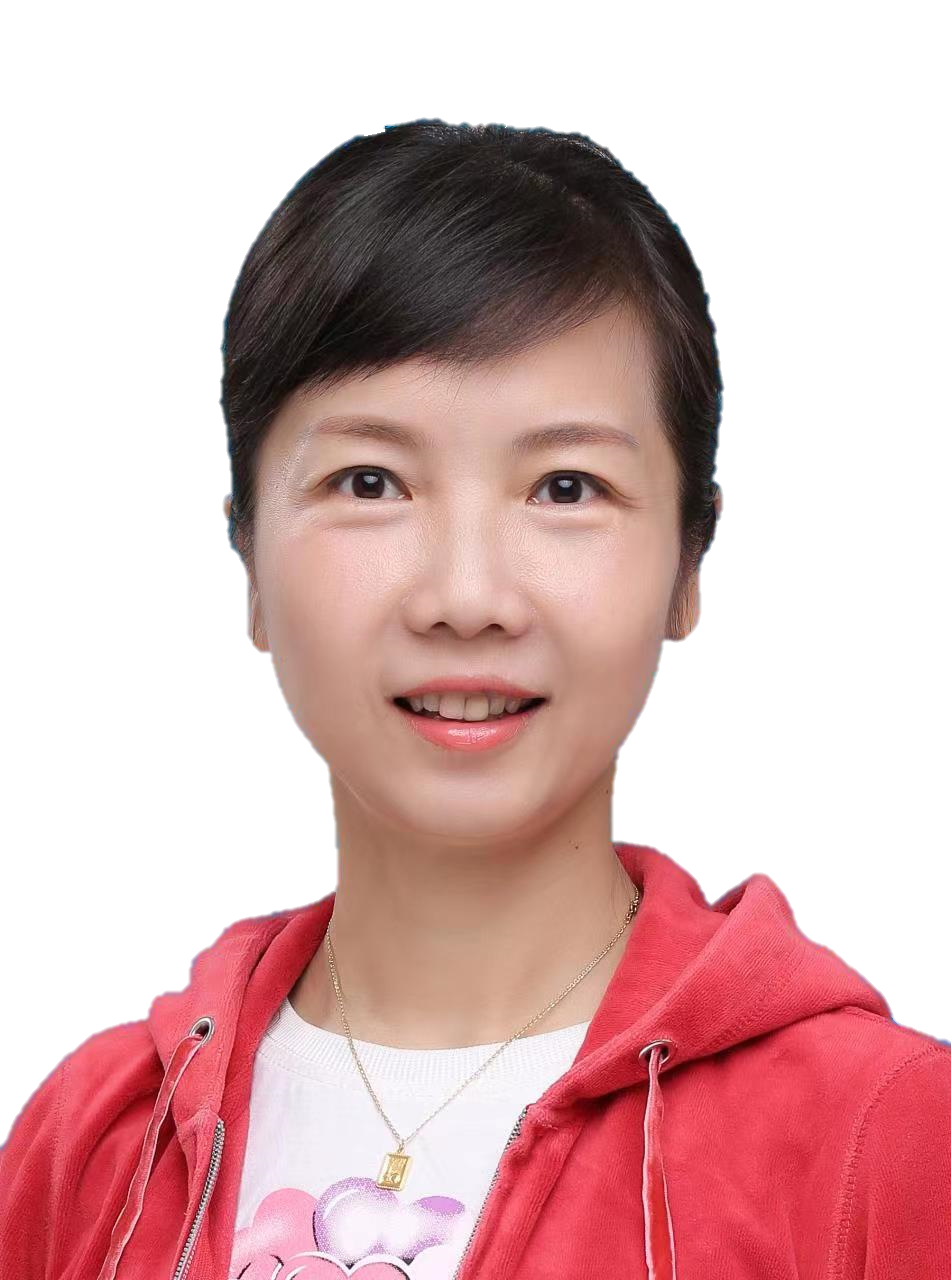}}]{Haifang Zhou} received her Ph.D. degree from National University of Defense Technology (NUDT), Changsha, China, in 2003.

She is now Professor of the College of Computer Science and Technology, NUDT. Her current research interests include bigdata processing and intelligent parallel computing. Dr. Zhou has published 60+ peer-reviewed papers, including those in highly regarded journals and conferences such as Concurrency and Computation Practice and Experience(CCPE), Journal of Software, Parallel Computing, International Conference on Parallel Processing(ICPP) etc.
\end{IEEEbiography}
\vspace{-40pt}

\begin{IEEEbiography}[{\includegraphics[width=1in,height=1.25in,clip,keepaspectratio]{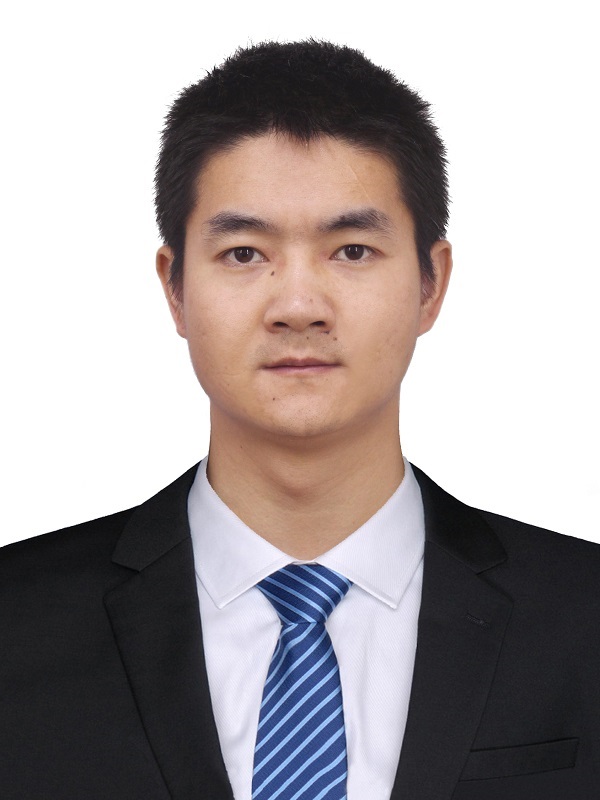}}]{Xinwang Liu} received his Ph.D. degree from National University of Defense Technology (NUDT), China, in 2013.

He is now Professor of College of Computer Science and Technology, NUDT. His current research interests include kernel learning and unsupervised feature learning. Dr. Liu has published 100+ peer-reviewed papers, including those in highly regarded journals and conferences such as IEEE T-PAMI, IEEE T-KDE, IEEE T-IP, IEEE T-NNLS, IEEE T-MM, IEEE T-IFS, ICML, NeurIPS, ICCV, CVPR, AAAI, IJCAI, etc. He serves as the associated editor of Information Fusion Journal. More information can be found at \url{https://xinwangliu.github.io/}.
\end{IEEEbiography}
% \begin{IEEEbiography}[{\includegraphics[width=1in,height=1.25in,clip,keepaspectratio]{fig1.png}}]{IEEE Publications Technology Team}
% In this paragraph you can place your educational, professional background and research and other interests.\end{IEEEbiography}

\end{document}